\documentclass{article}
\usepackage[T1]{fontenc}
\usepackage{iclr2027_conference,times}
\usepackage{amsmath,amsfonts}
\usepackage[hyphens]{url}
\usepackage{hyperref}
\usepackage{booktabs}
\usepackage{amssymb}
\usepackage{colortbl}
\usepackage{graphicx}
\usepackage{flafter}
\definecolor{PromptBorder}{HTML}{87A9C4}
\definecolor{PromptTint}{HTML}{EAF1F6}
\definecolor{JudgeBorder}{HTML}{C5A16C}
\definecolor{JudgeTint}{HTML}{F7F0E4}
\makeatletter
\newenvironment{appendixfigure}{%
  \par\addvspace{7pt}\noindent\begin{minipage}{\linewidth}%
  \centering\def\@captype{figure}%
}{\end{minipage}\par\addvspace{7pt}}
\newenvironment{appendixtable}{%
  \par\addvspace{7pt}\noindent\begin{minipage}{\linewidth}%
  \centering\def\@captype{table}%
}{\end{minipage}\par\addvspace{7pt}}
\makeatother
\newenvironment{caseblock}{%
  \par\addvspace{5pt}\noindent\begin{minipage}{\linewidth}%
}{\end{minipage}\par\addvspace{5pt}}
\newsavebox{\instructionbox}

\newcommand{\promptmeta}[3]{%
  {\fontsize{8.5}{10}\selectfont\color{black!70}%
  \textbf{Use:} #1\par\textbf{Inputs:} #2\par\textbf{Output:} #3\par}%
  \vspace{1pt}\hrule height .3pt\vspace{4pt}%
}
\newenvironment{promptbox}[2]{%
  \par\addvspace{4pt}\begingroup
  \def\instructioncolor{#2}%
  \setlength{\fboxrule}{.55pt}%
  \setlength{\fboxsep}{4pt}%
  \begin{lrbox}{\instructionbox}%
  \begin{minipage}{\dimexpr\linewidth-2\fboxsep-2\fboxrule\relax}
  \small\raggedright\setlength{\parindent}{0pt}\setlength{\parskip}{2pt}%
  \begingroup\setlength{\fboxsep}{2.5pt}%
  \colorbox{#2Tint}{\parbox{\dimexpr\linewidth-2\fboxsep\relax}{%
    \rmfamily\bfseries #1}}\par\endgroup\vspace{2pt}%
}{%
  \end{minipage}\end{lrbox}%
  \noindent\fcolorbox{\instructioncolor Border}{white}{\usebox{\instructionbox}}%
  \par\endgroup\addvspace{4pt}%
}

\title{Omni-InteractScope: Benchmarking\\
Proactive and Agentic Interaction in\\
Continuous Omnimodal Streams}

\newcommand{\oduauthor}[2]{\mbox{#1\textsuperscript{#2}}}
\newcommand{\authorrow}[1]{\noindent#1\par}
\newcommand{\paperAuthorRows}{%
  \authorrow{%
    \oduauthor{Yinsong Yan}{1,2},
    \oduauthor{Ling Wang}{1,2},
    \oduauthor{Yuxuan Wang}{2},
    \oduauthor{Yifan Yang}{2,3},
    \oduauthor{Ruiyang Xu}{2,3}%
  }%
  \authorrow{%
    \oduauthor{Junming Lin}{2,4},
    \oduauthor{Qi Chen}{2,3},
    \oduauthor{Meng Gao}{2,4},
    \oduauthor{Minghao Han}{2,5},
    \oduauthor{Yunfei Chu}{2}%
  }%
  \authorrow{%
    \oduauthor{Qize Yang}{2},
    \oduauthor{Xize Cheng}{2},
    \oduauthor{Jin Xu}{2},
    \oduauthor{Jibin Wu}{1,\textdagger}%
  }%
}
\newcommand{\paperAuthorBlock}{%
  \begin{minipage}[t]{\dimexpr\textwidth-2\tabcolsep\relax}
  \raggedright
  \setlength{\parskip}{0pt}
  \normalsize\bfseries
  \paperAuthorRows
  \vspace{4pt}
  \normalfont\normalsize
  \mbox{\textsuperscript{1}The Hong Kong Polytechnic University}\par
  \mbox{\textsuperscript{2}Alibaba Token Hub, Alibaba Group}\quad
  \mbox{\textsuperscript{3}Shanghai Jiaotong University}\par
  \mbox{\textsuperscript{4}Tsinghua University}\quad
  \mbox{\textsuperscript{5}Fudan University}\par
  \vspace{4pt}
  {\small Project page:
  \url{https://yinsongyan.github.io/Omni-InteractScope/}\par}
  \end{minipage}%
}
\author{\paperAuthorBlock}
\hypersetup{pdfauthor={Yinsong Yan, Ling Wang, Yuxuan Wang, Yifan Yang, Ruiyang Xu, Junming Lin, Qi Chen, Meng Gao, Minghao Han, Yunfei Chu, Qize Yang, Xize Cheng, Jin Xu, Jibin Wu}}

\iclrfinalcopy 

\begin{document}
\maketitle
\lhead{Preprint}
\begingroup
\renewcommand{\thefootnote}{\fnsymbol{footnote}}
\footnotetext[2]{Corresponding author.}
\endgroup

\begin{abstract}
Omnimodal large language models (Omni-LLMs) are increasingly expected to assist users while observing continuous audiovisual streams. However, existing evaluations provide limited insight into how reliably these models connect stream evidence and user needs to timely, sustained interventions. We introduce \textbf{Omni-InteractScope}, a benchmark of 1,479 human-annotated, cross-reviewed interactions with 1,790 temporal targets across eight task groups. It links input modalities, event triggers, intervention types, and user needs, covering event-timed tool invocation and query-free assistance alongside notifications, warnings, state tracking, and continuous responses. Our experimental results on the proposed benchmark reveal three critical limitations of current Omni-LLMs: (1) \textbf{Timely tool invocation remains unreliable}: all 12 evaluated models produce an on-time first response for at most 40\% of tool-invocation targets, and joint timing-and-tool accuracy reaches at most 32.0\%. (2) \textbf{Query-free reminders remain poorly covered}: all evaluated models achieve lower timely reference-target recall on query-free reminders than on explicit questions, measured on their respective target sets. Reminder recall never exceeds 23.0\%. (3) \textbf{End-to-end guidance coverage remains limited}: nine models cover none of the 26 guidance sequences in full, and the remaining three cover only one to three sequences each. Targeted controls further show that tool-invocation scores alone do not establish event grounding and that reminder coverage depends on response history and frequency. These findings motivate developing and evaluating streaming assistants for timely action, initiative under general user needs, and coverage of entire response sequences.
\end{abstract}

\begingroup
\setlength{\parskip}{5.6pt}
\setlength{\textfloatsep}{8pt plus 2pt minus 2pt}
\setlength{\floatsep}{8pt plus 2pt minus 2pt}
\setlength{\abovecaptionskip}{5pt}
\makeatletter
\renewcommand{\section}{\@startsection{section}{1}{\z@}%
  {-1.2ex plus -.3ex minus -.2ex}{.8ex plus .2ex minus .1ex}%
  {\large\sc\raggedright}}
\renewcommand{\subsection}{\@startsection{subsection}{2}{\z@}%
  {-1.3ex plus -.3ex minus -.2ex}{.5ex plus .1ex}%
  {\normalsize\sc\raggedright}}
\renewcommand{\paragraph}{\@startsection{paragraph}{4}{\z@}%
  {1ex plus .3ex minus .2ex}{-1em}{\normalsize\bf}}
\makeatother
\section{Introduction}
\label{sec:intro}
Omnimodal large language models (Omni-LLMs) such as GPT-4o, Qwen2.5-Omni, and Qwen3-Omni bring joint audiovisual understanding into conversational systems \citep{gpt4o2024,xu2025qwen25omni,xu2025qwen3omni}. Emerging streaming systems extend this ability to deciding when to speak, remain silent, or invoke tools \citep{chen2024videollmonline,wang2025mmduet2,seed2026realtime}. For an assistant observing a changing environment, these decisions must serve a user's ongoing needs: retrieve information when a requested event occurs, relay a relevant sound to a user with hearing impairment, or provide continue guidance as an activity unfolds (see Figure~\ref{fig:concept}). Useful assistance therefore requires connecting perceptual evidence to \emph{what intervention is needed, when it is useful, and whether further responses remain necessary}.

\begin{figure}[t]
\centering
\includegraphics[width=\linewidth]{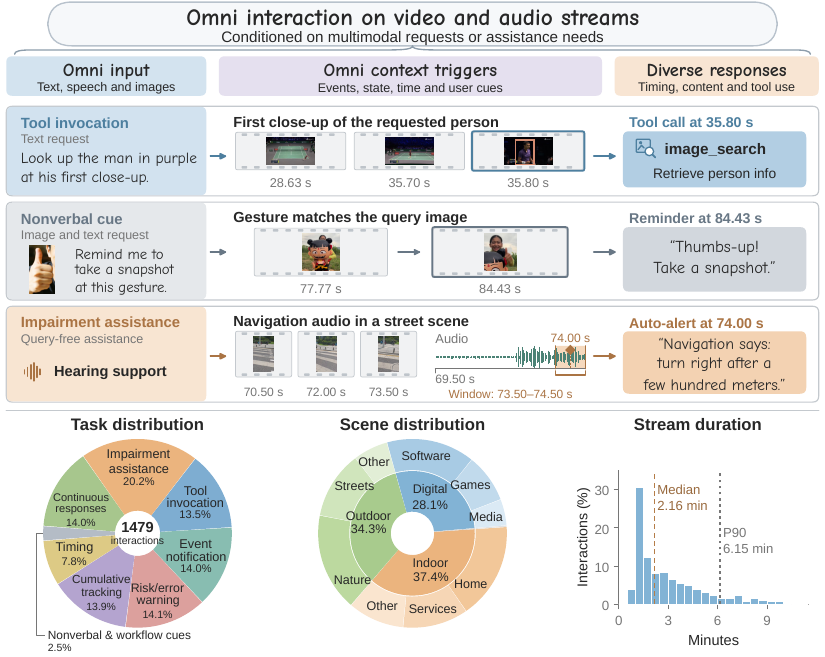}
\caption{\textbf{Overview of Omni-InteractScope.}
\textbf{Top:} Representative interactions connect user requests or general assistance needs with events in audiovisual streams and the content and timing of the required responses. The examples span event-timed tool invocation, a gesture reminder conditioned on an image and text request, and hearing
assistance under a general instruction without an event-specific query.
\textbf{Bottom:} Distributions of task types, scene categories, and stream durations across the benchmark's 1,479 interactions. Clips reused across multiple interactions are counted once for each interaction in which they appear.}
\label{fig:concept}
\end{figure}

Evaluation has progressed from video comprehension toward interactive assistance. Early works, such as Video-MME and LongVideoBench, assess understanding of supplied videos \citep{videomme2024,longvideobench2024}, while StreamingBench and OVO-Bench restrict observations to available stream prefixes and test responses to unfolding events \citep{lin2024streamingbench,niu2025ovo}. Recent benchmarks further evaluate multimodal turn-taking, audio-dependent proactivity, and event-triggered response windows \citep{omnimmi2025,zhao2026omnipro,omniinteract2026}. OmniAssistBench organizes goal-oriented assistance along predefined interaction paths \citep{sun2026omniassistbench}. StreamArena jointly evaluates perception, retrospection, proactivity, and tool utilization over hour-scale causal streams \citep{zhang2026streamarena}. Assistance without an event-specific question is also studied in VIABench \citep{liu2026viabench}. Table~\ref{tab:benchmark_comparison} summarizes this progression in terms of stream access and intervention objectives. Section~\ref{sec:related} provides a comprehensive discussion of related work.

An assistant's obligation extends beyond the content of its next answer. A specified event can require a tool call at a useful moment, a general assistance need can require intervention without an event-specific question, and a standing request can require further responses after an initial answer. Evaluating these requirements together calls for explicit links between user context, response content, valid timing, and successive targets. It also calls for distinguishing scene evidence from request priors and target coverage from the opportunities allowed to respond. This combination makes it possible to examine both whether an intervention obligation is fulfilled and which evaluation conditions shape the result.

We introduce \textbf{Omni-InteractScope}, a benchmark organized around the relationship between stream evidence, user needs, and required interventions. It contains 1,479 human-annotated, cross-reviewed interactions with 1,790 temporal targets. The assistance panel draws on 207 self-recorded first-person videos. The benchmark links four dimensions: input modalities, event triggers, intervention types, and user needs. Eight task groups span tool invocation, notifications, warnings, cumulative-state tracking, temporal conditions, nonverbal cues, continuous responses, and impairment assistance. Each target specifies response content and a valid time or interval. Interactions with multiple targets record the responses required after an initial answer. This organization supports evaluating both individual interventions and coverage of continuing obligations within a shared resource.

Within this broad coverage, we focus on two complementary demands of proactive interaction. Tool invocation requires selecting the appropriate tool and acting when a user-specified event occurs. Query-free assistance requires recognizing when a general user need warrants a reminder without an event-specific query. Paired controls examine the roles of scene evidence and response policy, while successive targets assess whether the assistant delivers the further responses required as an interaction unfolds.

Evaluation across 12 models reveals substantial limitations in both timely intervention and sustained assistance. Every evaluated model produces an on-time first response for at most 40\% of tool-invocation targets, and joint accuracy reaches at most 32.0\%. Timely reference-target recall on query-free reminders never exceeds 23.0\%. Although guidance-step recall reaches 51.7\%, nine models cover none of the 26 guidance sequences in full, and no model covers more than three.



Our main contributions are summarized as follows:
\begin{itemize}
\item We introduce Omni-InteractScope, a comprehensive benchmark for timely and sustained omnimodal intervention. It comprises 1,479 human-annotated, cross-reviewed interactions across eight task groups, with 1,790 targets specifying what to produce and when, including successive responses under continuing user needs.
\item We provide focused evaluation of event-timed action and query-free initiative through 200 tool-invocation targets and 122 query-free reminders. Our diagnostics distinguish tool choice from timing and test whether invocation scores depend on scene evidence. Paired controls further examine how explicit event cues, response scheduling, history, and response frequency affect coverage on the same
reminder targets.
\item We perform a comprehensive evaluation of 12 multimodal models, providing empirical insights into timing, query-free assistance, and sequence coverage, exposing gaps between tool selection and timely action, question answering, and query-free reminders, and step coverage and complete guidance.
\end{itemize}

\section{Related Work}
\label{sec:related}

\begin{table}[t]
\centering
\caption{\textbf{From video understanding to streaming intervention.}
Coverage of scene inputs and interaction objectives at benchmark level.}
\label{tab:benchmark_comparison}
\begingroup
\fontsize{8}{10}\selectfont
\setlength{\tabcolsep}{4.45pt}
\renewcommand{\arraystretch}{1.02}
\newcommand{\bcYes}{\textcolor[HTML]{28745C}{$\checkmark$}}
\newcommand{\bcNo}{\textcolor[gray]{0.55}{\textendash}}
\newcommand{\bcProbe}{\bcYes\textsuperscript{p}}
\resizebox{.96\linewidth}{!}{%
\begin{tabular}{lcrcccccc}
\toprule
\multicolumn{3}{c}{\textbf{Resource}} &
\multicolumn{2}{c}{\textbf{Evaluation}} &
\multicolumn{4}{c}{\textbf{Intervention objectives}} \\
\cmidrule(lr){1-3}\cmidrule(lr){4-5}\cmidrule(l){6-9}
\textbf{Benchmark} & \textbf{Modality} & \textbf{\#Videos} &
\shortstack{\textbf{Open-}\\\textbf{ended}} & \textbf{Streaming} &
\textbf{Proactive} & \textbf{Tool} &
\shortstack{\textbf{Query-}\\\textbf{free}} & \textbf{Successive} \\
\midrule
\multicolumn{9}{l}{\textit{(A) Offline video understanding}} \\
Video-MME \citep{videomme2024} & V, A, S & 900 & \bcNo & \bcNo & \bcNo & \bcNo & \bcNo & \bcNo \\
LongVideoBench \citep{longvideobench2024} & V, S & 3,763 & \bcNo & \bcNo & \bcNo & \bcNo & \bcNo & \bcNo \\
\midrule
\multicolumn{9}{l}{\textit{(B) Streaming understanding and interaction}} \\
StreamingBench \citep{lin2024streamingbench} & V, A & 900 & \bcNo & \bcYes & \bcProbe & \bcNo & \bcNo & \bcNo \\
OVO-Bench \citep{niu2025ovo} & V & 644 & Mix & \bcYes & \bcProbe & \bcNo & \bcNo & \bcYes \\
OmniMMI \citep{omnimmi2025} & V, A & 1,121 & Mix & \bcYes & \bcYes & \bcNo & \bcNo & \bcNo \\
OmniPro \citep{zhao2026omnipro} & V, A & 1,262 & \bcYes & \bcYes & \bcYes & \bcNo & \bcNo & \bcYes \\
IPIBench \citep{li2026ipibench} & V & 1,831 & \bcYes & \bcYes & \bcYes & \bcNo & \bcNo & \bcYes \\
OmniInteract \citep{omniinteract2026} & V, A & 250 & \bcYes & \bcYes & \bcYes & \bcNo & \bcNo & \bcYes \\
OmniAssistBench \citep{sun2026omniassistbench} & V, A & 300 & \bcYes & \bcYes & \bcYes & \bcNo & \bcNo & \bcYes \\
Omni-DuplexEval \citep{he2026omniduplexeval} & V, A & 660 & \bcYes & \bcYes & \bcYes & \bcNo & \bcNo & \bcYes \\
StreamArena \citep{zhang2026streamarena} & V, A & 243 & \bcYes & \bcYes & \bcYes & Exec. & \bcNo & \bcNo \\
\midrule
\multicolumn{9}{l}{\textit{(C) Assistance without an event-specific user query}} \\
VIABench \citep{liu2026viabench} & V & 761 & \bcYes & \bcYes & \bcYes & \bcNo & \bcYes & \bcYes \\
\midrule
\rowcolor[gray]{0.94}
\textbf{Omni-InteractScope (ours)} & \textbf{V, A} & \textbf{1,112}\textsuperscript{r} & \bcYes & \bcYes & \bcYes & Call & \bcYes & \bcYes \\
\bottomrule
\end{tabular}}
\vspace{3pt}

\parbox{.96\linewidth}{\fontsize{7.7}{9}\selectfont
V/A/S: video/audio/subtitles; queries are omitted. Audiovisual input does not establish audio necessity.
\bcNo: not an explicit objective; Mix: includes multiple-choice tasks.
Streaming includes causal prefix/replay, not necessarily native duplex inference;
\textsuperscript{p}proactivity uses repeated external probes.
Call: tool identity and timing; Exec.: execution and answer integration.
Query-free: no event-specific query. Successive: multiple targets under one continuing need.
\textsuperscript{r}Video records, including 207 self-recorded assistance videos.}
\endgroup
\end{table}

\paragraph{From video understanding to proactive interaction.}
Video-MME and LongVideoBench evaluate comprehension of supplied videos \citep{videomme2024,longvideobench2024}. StreamingBench and OVO-Bench restrict observations to available stream prefixes and test responses to unfolding events \citep{lin2024streamingbench,niu2025ovo}. OmniMMI and OmniPro extend evaluation to multimodal interaction and audio-dependent proactivity \citep{omnimmi2025,zhao2026omnipro}. IPIBench studies evolving requests, while OmniInteract evaluates turn-taking, interruption, and recovery \citep{li2026ipibench,omniinteract2026}. These works motivate causal evaluation of response content and timing. Omni-InteractScope records the intervention required by a user need, including targets that remain after an initial response.

\paragraph{Tool use and continuing response obligations.}
ToolBench and API-Bank evaluate tool-use plans and executable calls \citep{toolbench2024,apibank2023}. StreamArena brings tool utilization and proactivity into hour-scale causal streams, including external search and answer integration \citep{zhang2026streamarena}. Our event-timed tool invocation (T1) setting evaluates tool identity and invocation timing, with controls that distinguish request priors from scene evidence. Omni-DuplexEval includes continuous descriptions and event-conditioned reminders, and OmniAssistBench evaluates assistance along predefined paths \citep{he2026omniduplexeval,sun2026omniassistbench}. We retain successive targets within each interaction and report both individual-target and complete-guidance coverage. Table~\ref{tab:benchmark_comparison} separates these objectives from multiple independent questions about the same video.

\paragraph{Assistance under general user needs.}
VizWiz studies visual questions from blind users \citep{vizwiz2018}, while VIABench evaluates reminders, questions, and guided interaction \citep{liu2026viabench}. Vinci2 studies assistance without explicit prompts, and SoundWatch relays environmental sounds to deaf and hard-of-hearing users \citep{sitong2026vinci2,jain2020soundwatch}. Our assistance panel includes visual-
and hearing-assistance needs, distinguishing query-free reminders from explicit questions and multi-step guidance. Following concerns about response frequency in SPOT-Bench \citep{spot2026}, paired controls examine how history and generation opportunities affect reminder coverage and output volume on the same episodes.

\section{Omni-InteractScope}
\label{sec:data}
Omni-InteractScope comprises 1,479 interactions across eight task groups, organized around the interventions an assistant should deliver as a stream unfolds. Figure~\ref{fig:concept} summarizes the design, and Figure~\ref{fig:examples} illustrates each task group in details.

\begin{figure}[!t]
\centering
\includegraphics[width=.99\linewidth]{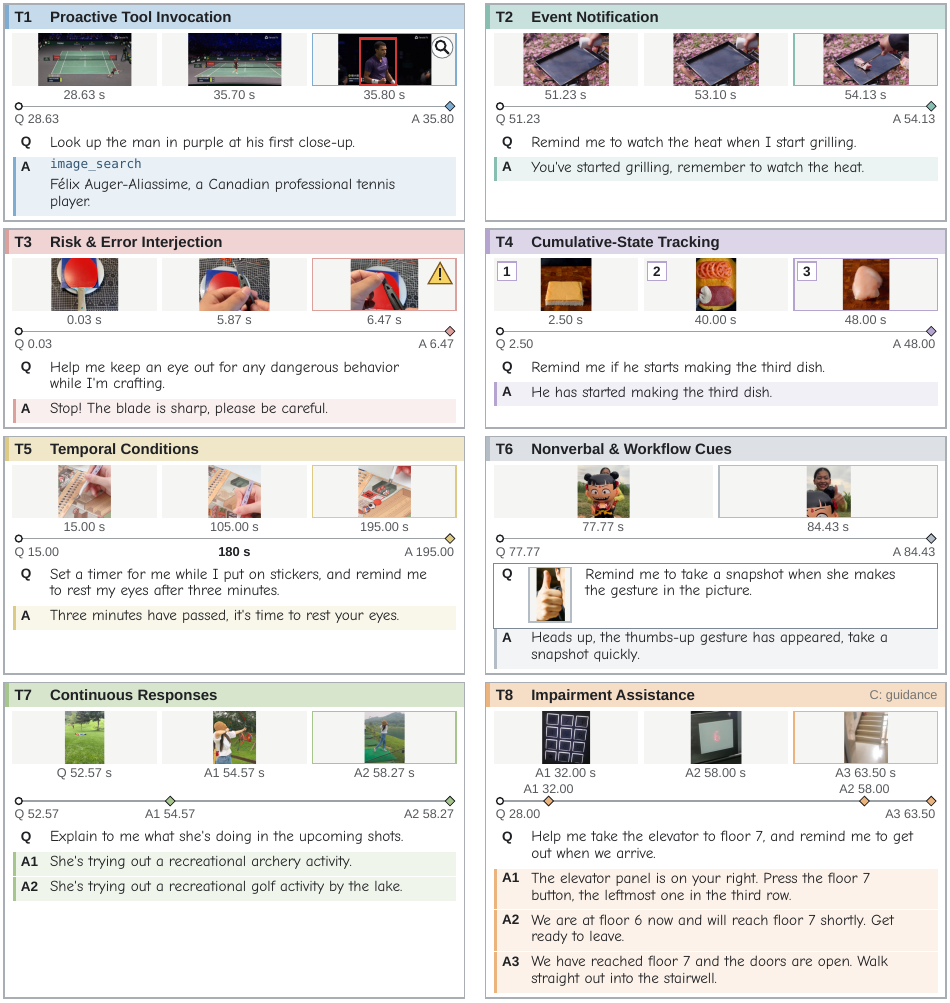}
\caption{\textbf{Representative interactions across the eight task groups of Omni-InteractScope.}
Each panel links a user request (Q), timestamped stream observations, and reference responses (A) to illustrate what intervention is required and when. The colored diamonds mark reference response times. T6 includes an image and text request, while T7 and the T8-C guidance example show how a single request can require successive responses as the stream unfolds.}
\label{fig:examples}
\end{figure}

\subsection{Designing for timely and sustained intervention}
The benchmark unit is an interaction: a video and audio stream, user context, and one or more reference interventions. Each intervention is a target specifying the required response and its reference time or valid interval. For a continuing request, the interaction retains successive targets in temporal order.

We organize this unit along four linked dimensions. \textbf{Input modalities} describe the available video, scene audio, and user context conveyed through text, speech, or image attachments. \textbf{Event triggers} describe the evidence or condition associated with a response, such as an object appearing, a sound, an accumulated count, or elapsed time. \textbf{Intervention types} specify what the assistant should produce: an answer, notification, warning, tool call, or successive commentary or guidance. \textbf{User needs} determine why that intervention matters, through an explicit request or a general visual- or hearing-assistance instruction.

Together, these dimensions connect stream evidence and user context to the interventions an assistant should provide. The eight task groups organize these requirements into a comprehensive evaluation of response content, timing, and coverage, encompassing both individual interventions and successive responses under continuing user needs.

\subsection{Task coverage and focal settings}
\paragraph{Responding to explicit requests.}
Seven task groups contain 1,180 interactions with different monitoring demands. \textbf{Event notification (T2)} asks the assistant to report a specified occurrence, while \textbf{risk/error interjection (T3)} calls for a warning when an unsafe or incorrect action occurs. \textbf{Nonverbal/workflow cues (T6)} connect a gesture or workflow event to the requested response, including matching a gesture to a query image. \textbf{Cumulative-state tracking (T4)} uses earlier observations to interpret current events. \textbf{Temporal conditions (T5)} introduce obligations governed by elapsed time. \textbf{Continuous responses (T7)} extend a standing request across successive commentary or translation segments.

\paragraph{Event-timed tool invocation.}
T1 tests whether a model invokes the appropriate tool when a specified event occurs. A request can indicate which tool to use before the stream establishes when to act. Each target therefore specifies a tool-name set and invocation time. Evaluation covers these requirements, with arguments and downstream execution outside its scope.

\paragraph{Assistance under explicit and general needs.}
The 299-interaction \textbf{impairment-assistance panel (T8)} covers visual- and hearing-assistance needs. \textbf{Query-free reminders (T8-A)} require identifying useful interventions under a general instruction, without an event-specific question. \textbf{Audiovisual questions (T8-B)} test answering explicit requests, while \textbf{multi-step guidance (T8-C)} tests successive responses along recorded trajectories. Both T7 and T8-C retain ordered targets within each interaction to represent continuing response obligations. Appendix~\ref{app:taxonomy} details the category boundaries.

\subsection{Data construction and annotation}
The benchmark combines 905 online video records for explicit-request tasks with 207 self-recorded first-person video records for impairment assistance. It spans indoor, outdoor, and digital-screen settings, with QA text approximately balanced between English and Chinese (744 and 735 interactions, respectively). The 1,479 interactions contain 1,790 targets, with multiple targets in 163 interactions. Appendix~\ref{app:construction} details task counts, scene and trigger distributions, duration statistics, and source provenance, including repeated source identities among video records.

Human annotators specified user context, reference responses, triggers, and response times or valid intervals. A separate group of human reviewers cross-checked the annotations to obtain the final benchmark. This review covers both response content and its temporal relation to the event. Appendix~\ref{app:annotation-audit} documents the supplemental consistency audit and review records.

\begingroup

\setlength{\textfloatsep}{8pt plus 2pt minus 2pt}
\section{Experiments}
\label{sec:results}
\subsection{Experimental setup}
\label{sec:experimental-setup}

\paragraph{Models.}
We evaluate 12 model configurations in three groups.
(1) \textit{Proprietary models} include Gemini-3.1-Pro \citep{deepmind2026gemini31pro} and Gemini-3.5-Flash \citep{deepmind2026gemini35flash}, and Qwen3.5-Omni-Plus \citep{team2026qwen3}.
(2) \textit{Open-source prompted MLLMs} include Qwen3-Omni
\citep{xu2025qwen3omni}, Qwen2.5-Omni \citep{xu2025qwen25omni},
Qwen3-VL \citep{bai2025qwen3vl}, and InternVL3.5 \citep{wang2025internvl35}.
(3) \textit{Streaming models} include LiveCC \citep{chen2025livecc},
MiniCPM-o 4.5 \citep{cui2026minicpmo45}, MMDuet2 \citep{wang2025mmduet2},
MOSS-VL \citep{wang2026mossvl}, and AURA \citep{lu2026aura}.

\paragraph{Response collection.}
Models receive causal inputs at query time, subsequent integer-second media times, and clip end; T8-A begins at stream start. The schedule is target-independent. \textit{Offline Probing} uses independent prefixes without prior responses. \textit{Online Rolling} replays pre-query observations with the request supplied, then retains native state across successive observations. Initialization outputs are excluded. Appendices~\ref{app:protocol} and~\ref{app:experimental-details} detail collection and adapters.

\paragraph{Scoring.}
Interaction $i$ contains $K_i$ targets, each specifying required content $y_{ik}$ and a valid time window $W_{ik}$. For response $j$ with content $\hat y_{ij}$ and timestamp $t_{ij}$, joint content--time eligibility is
\begin{equation}
E_{ikj}=\mathbb{I}[t_{ij}\in W_{ik}]\,C_i(\hat y_{ij},y_{ik}),
\label{eq:eligible-response}
\end{equation}
where $\mathbb{I}$ is the indicator function and $C_i$ returns 1 for correct content given the user context and 0 otherwise. T1 requires an exact match of tool-name sets, with no missing or extra tools. T2--T8 use a binary GPT-4o judge \citep{gpt4o2024} comparing the response with the request and reference, accepting equivalent wording. Timing is checked separately by the scorer.

T1--T7 and T8-B/C use a two-second tolerance around the reference time, with T8-B/C lower bounds clipped to query time. T8-A uses its annotated valid interval.
All timestamps use media time, excluding inference latency.
Appendix~\ref{app:protocol} specifies timestamp conventions and window boundaries.

\begin{table}[!t]
\centering
\caption{\textbf{Main results on explicit-request tasks (T1--T7).}
Joint content--timing accuracy (\%). Avg. equally weights T1--T4, T6, and T7.}
\label{tab:historical}
\begingroup
\fontsize{9}{9.4}\selectfont
\setlength{\tabcolsep}{2pt}
\renewcommand{\arraystretch}{1.02}
\begin{tabular*}{\linewidth}{@{\extracolsep{\fill}}lllrrrrrrrr@{}}
\toprule
\multicolumn{3}{c}{\textbf{Model configuration}} & \multicolumn{8}{c}{\textbf{Joint content--timing accuracy (\%) $\uparrow$}} \\
\cmidrule(lr){1-3}\cmidrule(l){4-11}
\textbf{Model} & \textbf{Size} & \textbf{Input} & \shortstack[c]{\textbf{T1}\\{\scriptsize Tool}} & \shortstack[c]{\textbf{T2}\\{\scriptsize Event}} & \shortstack[c]{\textbf{T3}\\{\scriptsize Risk}} & \shortstack[c]{\textbf{T4}\\{\scriptsize State}} & \shortstack[c]{\textbf{T5*}\\{\scriptsize Time}} & \shortstack[c]{\textbf{T6}\\{\scriptsize Cue}} & \shortstack[c]{\textbf{T7}\\{\scriptsize Cont.}} & \shortstack[c]{\textbf{Avg.}\\{\scriptsize 6 tasks}} \\
\multicolumn{3}{l}{\textit{Targets}} & 200 & 206 & 207 & 206 & 115 & 37 & 457 &  \\
\midrule
\rowcolor[gray]{0.96}
\multicolumn{11}{l}{\textit{Proprietary Models (prompted)}} \\
Gemini-3.1-Pro & N/R & V, A & 22.5 & 28.2 & 11.6 & \textbf{11.2} & N/A & \textbf{27.0} & 11.2 & \cellcolor[gray]{0.94}18.6 \\
Gemini-3.5-Flash & N/R & V, A & 29.5 & \textbf{31.6} & \textbf{14.0} & 10.2 & N/A & 24.3 & \textbf{11.4} & \cellcolor[gray]{0.94}\textbf{20.2} \\
Qwen3.5-Omni-Plus & N/R & V, A & \textbf{32.0} & 24.3 & 4.8 & 3.9 & N/A & 10.8 & 7.9 & \cellcolor[gray]{0.94}13.9 \\
\midrule
\rowcolor[gray]{0.96}
\multicolumn{11}{l}{\textit{Open-source MLLMs (prompted)}} \\
InternVL3.5 & 8B & V+ASR & 29.0 & 1.0 & 0.0 & 0.0 & N/A & 5.4 & 0.4 & \cellcolor[gray]{0.94}6.0 \\
Qwen3-VL & 8B & V+ASR & 19.5 & 4.9 & 0.5 & 0.0 & N/A & 2.7 & 3.1 & \cellcolor[gray]{0.94}5.1 \\
Qwen2.5-Omni & 7B & V, A & 11.0 & 0.5 & 0.5 & 0.5 & N/A & 0.0 & 1.5 & \cellcolor[gray]{0.94}2.3 \\
Qwen3-Omni & 30B-A3B & V, A & 26.5 & 2.9 & 1.4 & 1.0 & N/A & 2.7 & 5.7 & \cellcolor[gray]{0.94}6.7 \\
\midrule
\rowcolor[gray]{0.96}
\multicolumn{11}{l}{\textit{Streaming Models (rolling)}} \\
LiveCC & 7B & V & 0.0 & 1.5 & 0.5 & 0.0 & 0.0 & 0.0 & 0.7 & \cellcolor[gray]{0.94}0.4 \\
AURA & 8B & V & 0.0 & 4.9 & 0.5 & 1.5 & 0.0 & 2.7 & 0.7 & \cellcolor[gray]{0.94}1.7 \\
MMDuet2 & 3B & V & 0.0 & 3.4 & 1.0 & 0.5 & 0.0 & 2.7 & 1.8 & \cellcolor[gray]{0.94}1.6 \\
MOSS-VL & 11.3B & V & 1.5 & 9.2 & 1.9 & 5.8 & 0.9 & 8.1 & 5.9 & \cellcolor[gray]{0.94}5.4 \\
MiniCPM-o 4.5 & 9B & V, A & 1.5 & 8.3 & 1.4 & 1.0 & 1.7 & 2.7 & 1.1 & \cellcolor[gray]{0.94}2.7 \\
\bottomrule
\end{tabular*}
\endgroup
\par\vspace{3pt}
\begin{minipage}{\linewidth}\footnotesize
\textit{Notes.} Prompted models use \textit{Offline Probing} with independent causal-prefix calls,
while streaming models use \textit{Online Rolling} with retained state (Section~\ref{sec:experimental-setup}).
*T5 retains the earlier query-relative rolling evaluation as a separate diagnostic, excluded from every Avg. N/A denotes probing inputs without an explicit elapsed-time clock. V: video, A: native audio, ASR: scene transcripts.
A3B denotes active parameters, and N/R represents undisclosed size.
Bold marks column maxima in the six-task comparison.
\end{minipage}
\end{table}

T1--T7 score the first response assigned to each target. A premature or incorrect response cannot be rescued by a later correction. T8 instead maximizes content--time matches while preserving target and response order, using each response at most once. Unmatched targets and incomplete interactions score zero.

\paragraph{Metrics.}
Let $m_{ik}\in\{0,1\}$ indicate target success under these rules. For a reported
set of interactions $\mathcal{S}$, we aggregate target successes as
\begin{equation}
S(\mathcal{S})=
\frac{\sum_{i\in\mathcal{S}}\sum_{k=1}^{K_i}m_{ik}}
     {\sum_{i\in\mathcal{S}}K_i}.
\label{eq:target-score}
\end{equation}
We report $S$ as task accuracy for T1--T7 and timely reference-target recall for T8, in percentages. Unmatched T8 outputs incur no penalty, so recall does not measure selectivity or interruption cost.
Complete T8-C counts interactions with every guidance step covered, out of 26 sequences.
Appendix~\ref{app:scoring} gives formal metrics, content rubrics, and assignment rules.

\subsection{Main results}
\label{sec:main-results}
Tables~\ref{tab:historical} and~\ref{tab:assistance} report explicit-request and assistance results. Avg. equally weights T1--T4, T6, and T7, with T5 reported separately. Assistance results pool target recall and count complete guidance sequences. Configurations differ in inputs and execution interfaces (Appendix~\ref{app:protocol-control} examines within-model schedule effects).
After media-validity exclusions, evaluation covers 1,477 interactions and 1,788 targets (Appendix~\ref{app:evaluation-population}).

\paragraph{Task strengths vary across intervention types.}
Gemini-3.5-Flash leads the six-task Avg., while Qwen3.5-Omni-Plus leads tool invocation. Yet the best scores on cumulative-state tracking (T4) and continuous responses (T7) are only 11.2\% and 11.4\%, respectively. These weaknesses span both timing and content: Gemini-3.5-Flash responds prematurely on 172/206 T4 targets, while its T7 failures include 233/457 premature responses and 152/457 timely content failures (Appendix~\ref{app:failures}). Task-level and failure breakdowns therefore reveal limitations obscured by the aggregate ranking.

\begin{table}[!t]
\centering

\caption{\textbf{Main results on assistance tasks (T8).} Timely reference-target recall (\%) for query-free reminders (T8-A), audiovisual QA (T8-B), and guidance steps (T8-C), with 122/151/87 targets (360 overall). Overall pools hits across all targets. Complete T8-C counts fully covered guidance sequences out of 26. All configurations use full-clip causal collection, response-availability timestamps, and ordered matching (Section~\ref{sec:experimental-setup}). Bold marks column maxima.}
\label{tab:assistance}
\begingroup
\fontsize{9}{9.4}\selectfont
\setlength{\tabcolsep}{3pt}
\renewcommand{\arraystretch}{1.02}
\begin{tabular*}{\linewidth}{@{\extracolsep{\fill}}lllrrrrr@{}}
\toprule
\multicolumn{3}{c}{\textbf{Model configuration}} & \multicolumn{4}{c}{\textbf{Timely target recall (\%) $\uparrow$}} & \textbf{Coverage} \\
\cmidrule(lr){1-3}\cmidrule(lr){4-7}\cmidrule(l){8-8}
\textbf{Model} & \textbf{Size} & \textbf{Input} & \textbf{Overall} & \textbf{T8-A} & \textbf{T8-B} & \textbf{T8-C} & \textbf{Complete T8-C} \\
\midrule
\rowcolor[gray]{0.96}
\multicolumn{8}{l}{\textit{Proprietary Models}} \\
Gemini-3.1-Pro & N/R & V, A & \cellcolor[gray]{0.94}\textbf{46.39} & \textbf{22.95} & \textbf{62.25} & \textbf{51.72} & \textbf{3/26} \\
Gemini-3.5-Flash & N/R & V, A & \cellcolor[gray]{0.94}41.11 & 22.13 & 57.62 & 39.08 & 2/26 \\
Qwen3.5-Omni-Plus & N/R & V, A & \cellcolor[gray]{0.94}31.11 & 9.02 & 49.67 & 29.89 & 1/26 \\
\midrule
\rowcolor[gray]{0.96}
\multicolumn{8}{l}{\textit{Open-source MLLMs}} \\
InternVL3.5 & 8B & V+ASR & \cellcolor[gray]{0.94}14.17 & 0.00 & 26.49 & 12.64 & 0/26 \\
Qwen3-VL & 8B & V+ASR & \cellcolor[gray]{0.94}15.00 & 2.46 & 24.50 & 16.09 & 0/26 \\
Qwen2.5-Omni & 7B & V, A & \cellcolor[gray]{0.94}10.00 & 2.46 & 19.87 & 3.45 & 0/26 \\
Qwen3-Omni & 30B-A3B & V, A & \cellcolor[gray]{0.94}13.06 & 2.46 & 25.83 & 5.75 & 0/26 \\
\midrule
\rowcolor[gray]{0.96}
\multicolumn{8}{l}{\textit{Streaming Models}} \\
LiveCC & 7B & V & \cellcolor[gray]{0.94}3.61 & 0.00 & 8.61 & 0.00 & 0/26 \\
AURA & 8B & V & \cellcolor[gray]{0.94}6.94 & 0.00 & 15.89 & 1.15 & 0/26 \\
MMDuet2 & 3B & V & \cellcolor[gray]{0.94}1.94 & 0.00 & 4.64 & 0.00 & 0/26 \\
MOSS-VL & 11.3B & V & \cellcolor[gray]{0.94}3.33 & 0.00 & 7.28 & 1.15 & 0/26 \\
MiniCPM-o 4.5 & 9B & V, A & \cellcolor[gray]{0.94}3.06 & 0.00 & 7.28 & 0.00 & 0/26 \\
\bottomrule
\end{tabular*}
\endgroup
\end{table}

\paragraph{Reminder coverage is lower than question coverage.}
Every configuration covers a smaller fraction of query-free reminders (T8-A) than explicit questions (T8-B). Even Gemini-3.1-Pro, which leads both, reaches 23.0\% on reminders versus 62.3\% on questions. These are different event sets. Section~\ref{sec:initiative} tests guidance and response policies on matched reminder episodes.

\paragraph{Step coverage does not ensure complete guidance.}
Gemini-3.1-Pro covers 51.7\% of T8-C steps but completes only 3 of 26 guidance sequences. No open-source prompted or streaming configuration completes a sequence. Complete T8-C measures coverage of the annotated trajectory, not closed-loop user-task success. Section~\ref{sec:sequence-analysis} relates completion to individual-step coverage.

\section{Diagnostic Analysis}
\label{sec:analysis}
\subsection{Scene evidence and tool-invocation scores}
\label{sec:tools}
Figure~\ref{fig:tools}a shows that removing scene input does not lower Gemini-3.5-Flash's T1 accuracy. On the same 200 interactions, joint accuracy is 29.50\% with scene input and 33.00\% without it, with requests and the evaluation protocol held fixed. The paired difference is $-3.50$ percentage points.
Joint accuracy alone therefore does not establish event-grounded invocation. Appendices~\ref{app:scene-control} and~\ref{app:t1-tolerance} detail the paired ablation and temporal-tolerance analysis.

The request-only control reveals strong task priors. Gemini-3.5-Flash selects 96.5\% of tools from request text and public definitions alone. Emitting these choices after a fixed three-second delay gives 68.0\% joint success, the highest of four tested delays. This separate control uses a closed-label output interface and different response opportunities from the main evaluation, so it diagnoses request and timing priors rather than providing a matched model ranking (Appendix~\ref{app:temporal-priors}).

\begin{figure}[!t]
\centering
\setlength{\abovecaptionskip}{2pt}
\includegraphics[width=.99\linewidth]{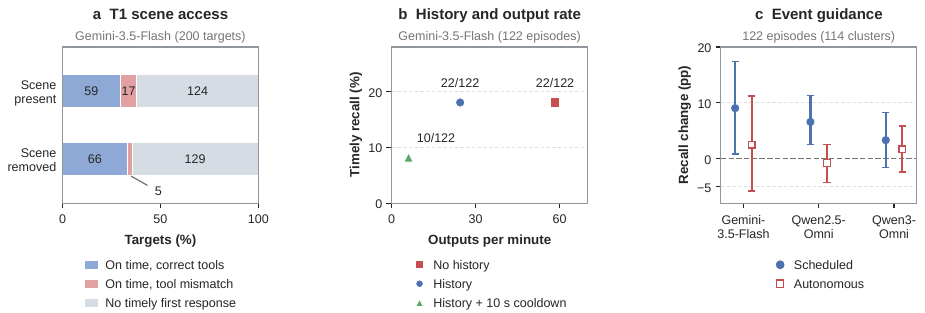}
\caption{\textbf{Scene access and response opportunities shape intervention scores.}
(a) T1 outcomes for Gemini-3.5-Flash with and without scene input (200 targets, two-second tolerance). Labels report target counts.
(b) T8-A timely recall versus output rate for Gemini-3.5-Flash with no history, the five most recent responses, or history plus a ten-second cooldown.
(c) Change in T8-A recall from general to event-specific instructions under scheduled and autonomous responding. Error bars show 95\% paired cluster-bootstrap confidence intervals.
Appendices~\ref{app:scene-control}, \ref{app:history-controls}, and~\ref{app:paired} detail the protocols and results.}
\label{fig:tools}
\end{figure}

\subsection{Reminder coverage under different response policies}
\label{sec:initiative}
As shown in Figure~\ref{fig:tools}b, giving Gemini-3.5-Flash response history reduces output volume by 58.1\%, while both conditions cover 22 of 122 T8-A targets. The matched targets change, with 12 gained and 12 lost. Adding a cooldown reduces outputs further but costs 9.84 percentage points of recall relative to history alone. The policy is implemented by skipping generation calls during the cooldown, restricting the opportunities to cover reminder targets (Appendix~\ref{app:history-controls}).

Figure~\ref{fig:tools}c compares event-specific and general instructions in separate runs on the same 122 T8-A episodes. Gemini-3.5-Flash and Qwen2.5-Omni benefit from event cues when a response is requested at an annotated valid time. Under autonomous responding, however, the effects are small and uncertain. The benefit of event cues is therefore clearer when the response time is supplied than when the model must decide when to intervene. Scheduling changes temporal information and response opportunities together, so this comparison does not isolate a planning mechanism.

\subsection{Complete guidance relative to step coverage}
\label{sec:sequence-analysis}
On the same 20 three-step T8-C episodes, Gemini-3.1-Pro covers 30 of 60 steps but completes only three sequences. Complete T8-C complements step recall by recording whether all required guidance was delivered.

Figure~\ref{fig:assistance} (Appendix~\ref{app:sequences}) compares observed completions with a reference based on success rates at each step position. An average of 2.4 complete sequences is expected if successes at different step positions are independent. The observed count is consistent with this reference and does not establish a separate memory or planning failure.
\endgroup

\section{Discussion}
\label{sec:discussion}
\paragraph{Invocation scores need evidence controls.}
The T1 controls show that request information and response timing can sustain joint success without scene inputs. A tool-and-time score should therefore be accompanied by evidence controls when used to assess event grounding. For streaming tool use \citep{zhang2026streamarena}, varying event occurrence and timing while holding requests fixed would test a stronger requirement than the scene ablation evaluated here.

\paragraph{Coverage depends on the response policy.}
Response history reduces Gemini-3.5-Flash's output volume with the same aggregate recall point estimate, while a cooldown reduces both outputs and coverage. These findings connect response-frequency evaluation \citep{spot2026} to the policies used during generation. Reporting history access and output rate alongside recall makes this dependence visible. Selectivity additionally requires evaluating the value and cost of interventions beyond the annotated targets, an important direction for future benchmark development.

\paragraph{Complete coverage has a separate measurement role.}
Users may need every guidance step even when individual responses are often correct. Complete T8-C expresses that requirement, while the sequence reference limits attribution of low completion to a separate memory or planning failure. For goal-oriented assistants \citep{sun2026omniassistbench}, explicit records of delivered guidance and pending obligations remain a design hypothesis to test through interventions on history and state.

\section{Conclusion}
\label{sec:conclusion}
We introduced Omni-InteractScope, a benchmark for timely and sustained intervention in continuous omnimodal streams. Its 1,479 human-annotated interactions span eight task groups, linking user needs and stream evidence to 1,790 targets that specify response content and timing. Evaluation across 12 models reveals substantial limitations in timely tool invocation, query-free reminders, and complete guidance coverage. These findings underscore the need to assess whether assistants fulfill the response requirements of an interaction, including interventions that remain necessary after an initial answer. Omni-InteractScope provides a shared basis for developing and evaluating assistants that respond to unfolding events and sustain support for continuing user needs.

\par\endgroup

\subsection*{AI use statement}
In this work, we used generative AI tools to help reorganize the benchmark task taxonomy, support translation, and improve the clarity and language of the manuscript. We also used an LLM judge for preliminary data validation prior to final human verification. Separately, LLM-based content judging was used in benchmark evaluation, as detailed in Appendix~\ref{app:scoring}. AI tools additionally assisted with organizing and presenting qualitative cases in Appendix~\ref{app:qualitative}.
Generative AI was not used to generate synthetic benchmark data, propose or refine research hypotheses, design the research methodology or experiments, implement the evaluated methods, or clean or reformat the benchmark dataset. It was also not used to develop theoretical models or new conceptual frameworks, formulate mathematical claims, or contribute to or write mathematical proofs. The authors retained responsibility for methodological decisions, qualitative judgments, and the scientific interpretation of the results. All benchmark media and annotations underwent rigorous human review. We take full responsibility for the final content of this work, including any text, claims, code, or other artifacts produced with the assistance of generative AI.

\subsection*{Reproducibility statement}
Sections~\ref{sec:data} and~\ref{sec:experimental-setup} describe benchmark design, data construction, the experimental setup, and the primary evaluation metrics. Appendix~\ref{app:construction} details dataset composition, taxonomy boundaries, annotation procedures, and consistency checks. Appendix~\ref{app:protocol} specifies observation schedules, timestamp conventions, temporal eligibility, response assignment, and uncertainty estimation, with worked examples of both matching rules. Appendix~\ref{app:experimental-details} provides input adapters, decoding configurations, and core prediction and judging prompts. Appendix~\ref{app:full-diagnostics} documents control conditions, sample sizes, and sensitivity analyses supporting the diagnostic findings. Appendix~\ref{app:qualitative} documents source frames, recorded model outputs, timing information, and scoring context for five illustrative cases.

\bibliography{references}
\bibliographystyle{iclr2027_conference}

\clearpage
\appendix
\raggedbottom
\renewcommand{\topfraction}{0.9}
\renewcommand{\bottomfraction}{0.8}
\renewcommand{\textfraction}{0.05}
\renewcommand{\floatpagefraction}{0.8}
\section*{Appendix Guide}
Appendix~\ref{app:construction} documents data, taxonomy, and annotation. Appendix~\ref{app:protocol} specifies observation and scoring rules, including worked examples. Appendix~\ref{app:experimental-details} gives model settings, adapters, and core prompts. Appendix~\ref{app:full-diagnostics} expands the three experimental questions and tests evaluation sensitivity.
Appendix~\ref{app:qualitative} examines actual model responses.

\section{Dataset, Taxonomy and Annotation}
\label{app:construction}
\subsection{Interaction records and task coverage}
The coverage statistics below expand the interaction and target representation in
Section~\ref{sec:data}, preserving successive targets within each interaction.

\begin{table}[!htbp]
\centering
\caption{Task coverage and annotated trigger modalities. V, A, and AV denote visual, auditory, and joint audiovisual triggers. Txt and Oth denote textual and other triggers. The five columns partition the reference targets within each task.}
\label{tab:coverage}
\begingroup
\fontsize{8}{10.2}\selectfont
\setlength{\tabcolsep}{3pt}
\renewcommand{\arraystretch}{1.12}
\begin{tabular*}{\linewidth}{@{\extracolsep{\fill}}llrrrrrrr@{}}
\toprule
\textbf{Task} & \textbf{Behavior} & \textbf{Interactions} & \textbf{Targets} & \textbf{V} & \textbf{A} & \textbf{AV} & \textbf{Txt} & \textbf{Oth} \\
\midrule
T1 & Tool invocation & 200 & 200 & 189 & 10 & 1 & 0 & 0 \\
T2 & Event notification & 207 & 207 & 150 & 36 & 18 & 3 & 0 \\
T3 & Risk/error interjection & 208 & 208 & 206 & 0 & 2 & 0 & 0 \\
T4 & Cumulative-state tracking & 206 & 206 & 171 & 3 & 6 & 3 & 23 \\
T5 & Temporal conditions & 115 & 115 & 36 & 0 & 0 & 0 & 79 \\
T6 & Nonverbal/workflow cues & 37 & 37 & 32 & 0 & 4 & 1 & 0 \\
T7 & Continuous responses & 207 & 457 & 374 & 62 & 16 & 5 & 0 \\
\midrule
T8-A & Query-free reminders & 122 & 122 & 101 & 16 & 5 & 0 & 0 \\
T8-B & Audiovisual QA & 151 & 151 & 135 & 11 & 5 & 0 & 0 \\
T8-C & Guided interaction & 26 & 87 & 79 & 7 & 1 & 0 & 0 \\
\midrule
\textbf{Total} &  & \textbf{1,479} & \textbf{1,790} & \textbf{1,473} & \textbf{145} & \textbf{58} & \textbf{12} & \textbf{102} \\
\bottomrule
\end{tabular*}
\endgroup
\end{table}
Table~\ref{tab:coverage} expands Figure~\ref{fig:concept}. T1--T7 contain 1,180
interactions and 1,430 targets, and T8 contains 299 interactions and 360 targets.
Multiple targets occur in 163 interactions. T1 has one target per interaction,
whereas T7 has 457 targets in 207 interactions and T8-C has 87 in 26.

\paragraph{Evaluation population.}
\label{app:evaluation-population}
The benchmark contains 1,479 interactions, 1,790 targets, and 1,112 video records. Evaluation uses 1,477 interactions and 1,788 targets: 1,063 interactions and 1,313 targets in T1--T4, T6, and T7, 115 in the separate T5 diagnostic, and 299 T8 interactions with 360 targets. Incomplete model trajectories retain their targets in these denominators (Appendix~\ref{app:incomplete-interactions}).

Request formats are text (1,202 interactions), audio (124), text with an image (25), text with audio (6), and no event-specific query (122, exactly T8-A).
Request modality and trigger modality describe different parts of the interaction: a visual event can, for example, indicate when to search the current audio. Actual model inputs are listed in Table~\ref{tab:adapter-inputs}.

\subsection{Taxonomy definitions and boundaries}
\label{app:taxonomy}
Trigger labels describe annotated evidence rather than establishing that a
modality is indispensable under every input condition. Intervention type describes
the output, while user need explains its purpose. Similar outputs can therefore
arise from explicit event requests or general assistance instructions, with
different requirements for initiative.

\paragraph{Requested actions and notifications (T1--T3).}
T1 requires invoking the reference tool set when the event occurs; tool identity
and timing are scored, while arguments and execution are outside the metric.
T2 reports a requested occurrence. T3 additionally requires communicating an
unsafe or incorrect action: merely announcing the event does not convey the
required warning.

\paragraph{What makes a response due (T4--T6).}
T4 depends on accumulated observations, such as the start of a third dish after
two earlier preparations. T5 depends on an elapsed duration and is reported
separately. T6 responds to a gesture or workflow cue, which can require matching
the observed gesture to a query image. A query attachment specifies the sought
cue; the stream supplies evidence that it has occurred.

\paragraph{General assistance and explicit questions (T8-A/B).}
T8-A supplies a general visual- or hearing-assistance need without an
event-specific question. The model must identify a useful reminder from the
stream. T8-B instead names the immediate information need through an explicit
audiovisual request. Similar response wording can therefore reflect different
requirements for initiative.

\paragraph{Continuing responses and actionable guidance (T7, T8-C).}
Both contain successive targets under one standing need. T7 covers continuing
commentary or translation. T8-C requires actionable instructions justified by
successive states along a recorded assistance trajectory and defines the unit
for complete-guidance reporting. T1--T7 group interventions and triggering
conditions, whereas T8 groups assistance needs; timely action and sustained
coverage can consequently apply to several task groups.

\subsection{Sources and annotation}
The explicit-request panel contains 905 video records: 888 from Douyin, 11 from
Xiaohongshu, and six from Bilibili, with 495 first-person and 410 third-person
records. The assistance panel uses 207 self-recorded first-person videos for
261 visual-impairment and 38 hearing-impairment interactions. QA text is approximately balanced
between English and Chinese at 744 and 735 interactions, respectively; scene audio retains its
original language. Indoor, outdoor, and digital-screen settings account for
37.4\%, 34.3\%, and 28.1\% of interactions. Median clip duration is 2.16 minutes,
and the 90th percentile is 6.15 minutes. Figure~\ref{fig:concept} counts each
interaction in these distributions and uses 30-second duration bins.

The 1,112 video records are storage and analysis units, not necessarily distinct
original sources. Seven groups in the explicit-request panel share platform
identities. Uncertainty estimates use video-record clusters; the source inventory
and identifier discrepancies are documented in the provenance materials.

Construction proceeds from source records to interaction annotations and then
cross-review. For each interaction, annotators record the user request or general
need, query time and attachments where applicable, and the required interventions.
Each target retains its content, triggering evidence, and time or interval. Related
targets remain ordered in one interaction, preserving the distinction between a
continuing need and independent questions. A separate group of human reviewers
cross-checks the annotations to produce the final benchmark.

\subsection{Annotation consistency checks}
\label{app:annotation-audit}
A stratified annotation-metadata audit conducted before release filtering contains 64 primary reviews, split evenly
between T1 and T8-A, and 16 secondary reviews. Sampling covers less frequent tools
and trigger modalities without using evaluated-model predictions. Final records
mark evidence support positive in 62/64 cases, timing in 63/64, modality in 62/64,
and tool or assistance need in 63/64. Five cases include documented
Gemini-3.5-Flash-assisted review with human confirmation.

A separate trigger-label check used Gemini-3.5-Flash and Gemini-3.1-Pro on six
targets from four videos, providing both models with the same video crops,
original audio, and annotation context. Both assigned all six triggers to the
visual category. These labels are incorporated into Table~\ref{tab:coverage} and
the modality analysis; reference content and timing are unchanged. Individual
judgments, adjudication notes, and revision records are retained with the audit.

\subsection{From scene evidence to an annotation record}
Figure~\ref{fig:examples} pairs source frames with requests and reference responses,
including accumulated state, elapsed time, query-image cues, and successive targets.
Figure~\ref{fig:concept} additionally illustrates query-free hearing assistance
with original scene audio represented by its measured energy envelope.
Figure~\ref{fig:annotation-record} expands the cumulative-state example to show
how the request, evidence, target time, and response are represented. The
two-second tolerance is applied by the evaluator to a point annotation; it is
not an additional human-annotated interval.

\begin{appendixfigure}
\centering
\includegraphics[width=\linewidth]{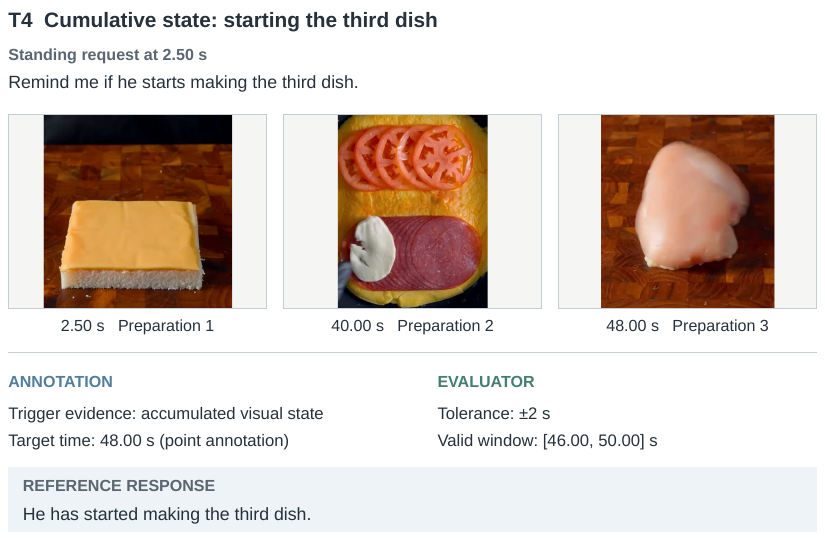}
\caption{\textbf{An annotated cumulative-state interaction (T4).}
Three source frames show successive dish preparations. The request and reference
response are reproduced from the annotation. Count labels identify the illustrated
preparations, not model predictions. The evaluator derives $[46,50]$ seconds
from the 48-second point target. The frame strip provides selected evidence;
the model observes the causal stream rather than this summary.}
\label{fig:annotation-record}
\end{appendixfigure}

\section{Evaluation Protocol and Scoring}
\label{app:protocol}
\label{app:adapters}
\label{app:scoring}
\subsection{Observation schedules and response history}
\paragraph{Main decision grid.}
For T1--T4, T6, T7, and T8, decisions occur at
$\{q,D\}\cup\{k\in\mathbb{Z}:q\leq k\leq D\}$, where $q$ is query time and
$D$ is clip duration, restricted to available media. A noninteger query time is
included before the next integer tick. T8-A starts at stream onset, and T8
decisions require an available video frame. Models receive no reference answers,
target times, or annotation-derived end times. The scheduled T8 controls are the
explicit exception: they request one response at an annotated valid time
(Appendix~\ref{app:paired}).

Prompted models make independent prefix calls without previous responses.
Stateful adapters retain a session, initialized at query arrival by replaying
earlier observations with the request already supplied. Initialization outputs
are excluded from evaluation; native state, including generated history, is
retained. Thus the main comparison shares decision opportunities and access to
prior observations, while input representation and memory mechanisms differ.
The additional Gemini-3.5-Flash history controls use their own regenerated baseline
(Appendix~\ref{app:history-controls}).

Execution may stop after the final target deadline only when earlier outputs
are invariant to stopping. All earlier decision opportunities are retained.
MiniCPM-o 4.5 and MOSS-VL run in full, and LiveCC must cover the entire last response
group.

\paragraph{Separate T5 diagnostic.}
T5 retains 115 rolling interactions beginning at query time without historical
initialization, with observations at $q,q+1,q+2,\ldots$ and adapter onset
timestamps. These results use a different schedule from the main six-task
comparison. Prompted T5 inputs lack an explicit elapsed-time clock, so their
cells are N/A. Every model's Avg. excludes T5.

\paragraph{Response units and timestamps.}
All times refer to the media clock. For T1--T4, T6, and T7, prompted responses
use prefix decision time. MiniCPM-o 4.5 and MOSS-VL use saved turn or sentence onset;
MMDuet2 and AURA return within one tick, making onset and collection time equal.
LiveCC chunks are assembled within two seconds of the first post-query chunk
and credited at that chunk's time, without chaining consecutive gaps.
Responses with saved onset before the query are excluded. For T8, responses use
collection time; LiveCC continuations are joined at the utterance boundary or
stream end and credited at the last chunk. This measures when assistance content
becomes available, rather than wall-clock serving latency.

\subsection{Notation and temporal eligibility}
\label{app:metric-definitions}
We record response $j$ in interaction $i$ as $(t_{ij},\hat y_{ij})$, with a
media timestamp and response content. Interaction $i$ contains $K_i$ reference
targets. Each target $k$ specifies required content $y_{ik}$ and an acceptable interval $W_{ik}$.
Let $C_i(\hat y,y_{ik})\in\{0,1\}$ denote content
correctness given the user context. Content and time are checked separately.

T1--T7 accept timestamps in $W_{ik}=[\tau_{ik}-2,\tau_{ik}+2]$ seconds around the
annotated target time $\tau_{ik}$. T8-A uses the annotated valid interval without
extension. T8-B and T8-C use $W_{ik}=[\max(q_i,\tau_{ik}-2),\tau_{ik}+2]$, where $q_i$ is
query time. All endpoints are inclusive. Equation~\ref{eq:eligible-response}
defines $E_{ikj}$ as joint content--time eligibility, using the indicator
function $\mathbb{I}[\cdot]$. Output timestamps follow the
response units and media-time conventions in Appendix~\ref{app:adapters}.

\subsection{Content criteria and judge inputs}
T1 compares the predicted and reference tool-name sets, accepting supported formatting variations while requiring no missing or extra tools. 

For T2--T8, the content judge is GPT-4o, with temperature
zero and a ten-token verdict budget. It receives the request, reference, and
candidate response, without scene media or timestamps. T8 adds the target user
group and subtask to the request context. Timing is checked separately by the
scorer. Table~\ref{tab:content-rubrics} summarizes the criteria used by the recorded
prompts, following rubric-based LLM evaluation \citep{llmjudge2023,geval2023}.

\begin{table}[!htbp]
\centering
\caption{Content criteria. Equivalent wording is accepted. Each rubric evaluates
correspondence to the reference target, independently of response timing.}
\label{tab:content-rubrics}
\begingroup
\fontsize{9}{10.2}\selectfont
\setlength{\tabcolsep}{3pt}
\renewcommand{\arraystretch}{1.12}
\begin{tabular*}{\linewidth}{@{\extracolsep{\fill}}p{.16\linewidth}p{.37\linewidth}p{.40\linewidth}@{}}
\toprule
\textbf{Rubric} & \textbf{Required content} & \textbf{Failure examples} \\
\midrule
General & Correct facts, names, quantities, and yes/no decisions.
& Wrong facts, hallucinations, or refusals. \\
Event & The same event with its key subject, object, and action.
& Generic narration, an unrelated event, or an incorrect participant or action. \\
Translation & The reference meaning, allowing synonyms and differences in fluency.
& Omitted meaning or added content. \\
Assistance & The target information and any essential action. A guidance step
must state the currently actionable instruction.
& Contradictory directions, unsafe advice, invented critical details, a generic
warning, or repetition of the user's intention. \\
\bottomrule
\end{tabular*}
\endgroup
\end{table}

Task and annotated language determine the general, event, or translation rubric; T8 uses the assistance rubric. The main comparison and paired controls require a valid Correct or Incorrect verdict. 

The judge assesses compatibility with the textual reference. Equivalent content in another language is accepted, so language-group results do not measure language compliance. The exact assistance prompt appears in Appendix~\ref{app:core-prompts}.

\paragraph{Incomplete interactions.}
\label{app:incomplete-interactions}
Each Gemini configuration returns complete traces for 1,061 of 1,063 interactions.
The remaining two interactions each contain one target, in T3 and T6.
These targets count as unsuccessful, retaining all 1,313 targets in the evaluation.

\subsection{First-response assignment for T1--T7}
All T1--T7 targets are point annotations. For the six-task main comparison,
streaming outputs first receive their onset timestamps, with initialization
responses removed as specified in Appendix~\ref{app:adapters}. LiveCC chunks
within two seconds of a group's first post-query chunk are assembled before
assignment. Trailing continuation markers are removed and immediately repeated
identical chunks within a group are counted once. Repeats in later groups remain
separate responses. A group is not extended beyond two seconds to complete its
meaning, and it is not discarded simply because its last chunk has a continuation
marker. This avoids interpreting prolonged commentary as silence.

Outputs are then sorted by timestamp and
merged before assignment: keep the first output, discard outputs within two
seconds of its timestamp, then repeat from the next remaining output. Each group
retains only its first text and timestamp. Grouping is anchored to the retained
output, not chained through consecutive gaps.

For one interaction, order target times as $\tau_1\leq\cdots\leq\tau_K$ and let
$U_k$ contain outputs consumed by earlier targets, initially empty. Define
\begin{equation}
b_1=-\infty,\qquad b_k=\frac{\tau_{k-1}+\tau_k}{2}\ (k>1),\qquad
J_k=\{j\notin U_k:b_k<t_j\leq\tau_k+2\}.
\end{equation}
If $J_k$ is empty, target $k$ fails. Otherwise it consumes the earliest output
$j_k\in J_k$, preserving recorded order for timestamp ties, and succeeds if
\begin{equation}
m_k=\mathbb{I}[t_{j_k}\geq\tau_k-2]C(\hat y_{j_k},y_k).
\end{equation}
A selected response is consumed even if premature or incorrect, so a later
correction cannot rescue that target. The midpoint lower bound is strict and
the two-second tolerance endpoints are inclusive. The rule applies to every
T7 target, including those after an earlier failed response.

\subsection{Ordered matching for assistance}
Only temporally eligible pairs under the windows in
Appendix~\ref{app:metric-definitions} are sent to the judge.
For a fixed interaction, let $E_{kj}$ denote the joint content--time eligibility
in Eq.~\ref{eq:eligible-response}, omitting the interaction index.

Matching maximizes the number of eligible pairs while preserving target and
output order. For the suffix beginning at target $k$ and output $j$,
\begin{equation}
D_{kj}=\max\{D_{k+1,j},\ D_{k,j+1},\ E_{kj}(1+D_{k+1,j+1})\},
\end{equation}
with zero-valued boundaries. An eligible diagonal is preferred when it achieves
the optimum. Otherwise, a target is skipped only if this gives a strictly higher
suffix value than skipping an output. One output can cover at most one target.
Unlike T1--T7, a later eligible response can match after an earlier failed attempt.
Target recall and complete-sequence coverage then follow
Eqs.~\ref{eq:target-score} and~\ref{eq:sequence-coverage}.

Unmatched responses incur no penalty. Appending outputs while retaining the
existing outputs and their order cannot reduce the optimal number of matches.
Timely reference-target recall therefore measures coverage, without measuring
false-alarm rate or intervention cost. References do not enumerate every useful
intervention, so an unmatched response cannot automatically be labeled invalid.
We therefore report output volume alongside coverage in
Appendix~\ref{app:history-controls}.

For LiveCC, continuation chunks ending in an ellipsis are concatenated until
the next completed chunk and credited at that last chunk's media time. An
unfinished final fragment remains at its last available time and is marked
incomplete. Other adapters retain their own output boundaries. This processing
uses no target information and never credits later content at utterance onset.

\subsection{Task aggregates and complete-sequence coverage}
Let $m_{ik}\in\{0,1\}$ indicate target success under the applicable rule, and let
$\mathcal{S}$ be the set of interactions being reported. The target score
$S(\mathcal{S})$ in Eq.~\ref{eq:target-score} divides total successes by total
targets. We report this as task accuracy for T1--T7 and target recall for T8.
Table~\ref{tab:historical}'s Avg. equally weights the six task accuracies for
T1--T4, T6, and T7. Table~\ref{tab:assistance}'s Overall pools all 361 targets
across T8-A, T8-B, and T8-C rather than averaging the three subtask recalls.
T5 is retained as a separate rolling diagnostic and marked N/A for prompted
inputs, which omit an explicit elapsed-time clock.
In T7 and T8-C, an interaction with two successful targets out of three contributes
two successes and three targets to this score.

For successive guidance in T8-C, we also report the fraction of interactions in
which every reference step is covered:
\begin{equation}
C_{\mathrm{sequence}}(\mathcal{S})=
\frac{1}{|\mathcal{S}|}\sum_{i\in\mathcal{S}}
\mathbb{I}\!\left[\sum_{k=1}^{K_i}m_{ik}=K_i\right].
\label{eq:sequence-coverage}
\end{equation}
Table~\ref{tab:assistance} reports the complete-sequence numerator as Complete T8-C,
out of 26 interactions. This distinguishes isolated correct steps from coverage of the entire
annotated sequence; it does not measure closed-loop task completion.

\subsection{Paired uncertainty estimates}
\label{app:uncertainty}
Unless otherwise stated, reported differences use 95\% percentile confidence
intervals from 10,000 bootstrap resamples of video-record clusters. Each resample
retains all targets and paired conditions within a selected record. We recompute
the stated aggregate, preserving task weighting where applicable. Differences
are in percentage points. These exploratory intervals have no multiplicity
correction and quantify sampling uncertainty, excluding judge error and API
run-to-run variation. A zero-width interval reflects constant observed paired
differences, not population equivalence. The complete-guidance permutation
reference in Appendix~\ref{app:sequences} is defined separately.
\subsection{Worked examples of response assignment}
\label{app:walkthrough}
Figure~\ref{fig:matching-walkthrough} contrasts the two assignment rules using
constructed examples. Content verdicts are stipulated for illustration;
Appendix~\ref{app:qualitative} shows real model outputs and stored judgments.

\begin{appendixfigure}
\centering
\includegraphics[width=\linewidth]{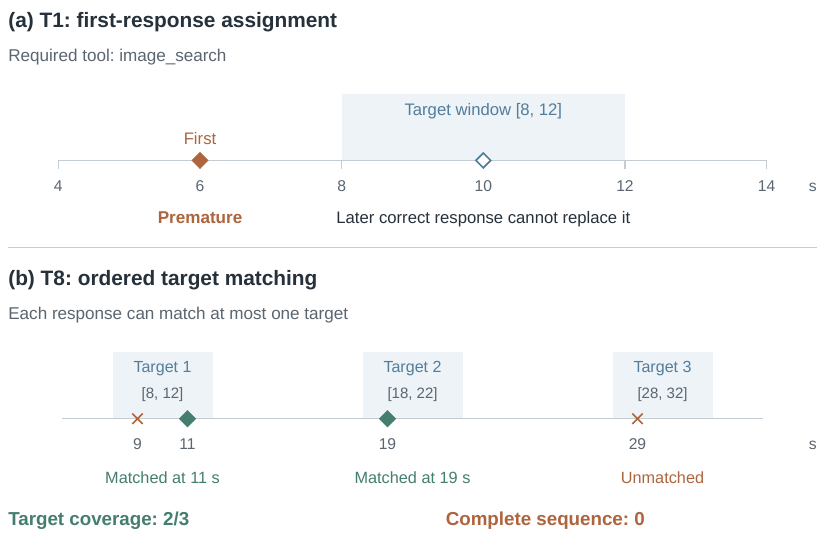}
\caption{\textbf{Timing, content, and assignment play separate roles.}
(a) T1 assigns the first response, even when a later response would be eligible.
The 6- and 10-second outputs are in separate groups because they are more than
two seconds apart. (b) T8 can match a later eligible response after an incorrect
attempt. Shaded intervals are inclusive valid windows. Diamonds mark correct
content and crosses mark incorrect content; a correct response can still fail
timing or remain unassigned. These are constructed scoring examples.}
\label{fig:matching-walkthrough}
\end{appendixfigure}

\paragraph{First-response assignment (T1--T7).}
The target requires \texttt{image\_search} at 10 seconds. The model emits that
tool at 6 seconds and again at 10 seconds. The first response precedes the
$[8,12]$ window and determines failure, giving $m_{i1}=0$; the later response
cannot replace it. Conversely, a timely \texttt{web\_search} would fail tool
identity. Both content and timing are necessary.

\paragraph{Ordered matching (T8).}
The first target matches at 11 seconds after an incorrect 9-second attempt,
and the second matches at 19 seconds. The third has only an incorrect candidate,
giving $2/3$ target coverage and zero complete sequences. If an eligible response
at 31 seconds also covers the third target, coverage becomes $3/3$ and the
sequence is complete. Each response can match at most one target. Unmatched
outputs do not reduce recall and are not automatically false positives, which
is why coverage is reported alongside output volume.

\section{Experimental Details}
\label{app:experimental-details}
\subsection{Input modalities and context}
\label{app:input-details}
At decision time $t$, prompted models receive user context and video frames
sampled from the available prefix $[0,t]$ at 1 fps, capped at 32 frames.
Frame presentation timestamps never exceed $t$. Native scene audio uses the
corresponding available prefix or, for streaming adapters, successive causal chunks.
V+ASR uses only completed transcript segments under the declared audio-dependency
and context policy. It does not necessarily transcribe the whole prefix or
represent non-speech sounds. Query attachments follow the supported input path.
Native adapters retain their own context windows.

\begin{table}[!htbp]
\centering
\caption{Assistance input paths. ``Text'' in the last column means the
annotated query text without consuming the separate query-audio attachment.
Every configuration also receives the annotated text.}
\label{tab:adapter-inputs}
\begingroup
\fontsize{9}{10.2}\selectfont
\setlength{\tabcolsep}{3pt}
\renewcommand{\arraystretch}{1.12}
\begin{tabular*}{\linewidth}{@{\extracolsep{\fill}}llll@{}}
\toprule
\textbf{Model} & \textbf{Interface} & \textbf{Scene audio} & \shortstack[l]{\textbf{Spoken-query}\\\textbf{attachment}} \\
\midrule
\multicolumn{4}{l}{\textbf{Proprietary Models}} \\
Gemini-3.1-Pro & Prefix & Native prefix & Audio and text \\
Gemini-3.5-Flash & Prefix & Native prefix & Audio and text \\
Qwen3.5-Omni-Plus & Prefix & Native prefix & Audio and text \\
\midrule
\multicolumn{4}{l}{\textbf{Open-source MLLMs}} \\
Qwen3-Omni & Prefix & Native prefix & Audio and text \\
Qwen2.5-Omni & Prefix & Native prefix & Audio and text \\
Qwen3-VL & Prefix & Completed-chunk ASR & Text \\
InternVL3.5 & Prefix & Completed-chunk ASR & Text \\
\midrule
\multicolumn{4}{l}{\textbf{Streaming Models}} \\
LiveCC & Stateful & None & Text \\
MMDuet2 & Stateful & None & Text \\
MOSS-VL & Stateful & None & Text \\
MiniCPM-o 4.5 & Stateful & Native chunks & Text \\
AURA & Rolling video & None & Text \\
\bottomrule
\end{tabular*}
\endgroup
\end{table}
All T8 configurations receive annotated query text or the general assistance
instruction for T8-A. Table~\ref{tab:adapter-inputs} distinguishes scene audio
from a separate spoken-query attachment. V+ASR assistance uses completed
independent one-second audio chunks. All 85 T8 spoken-query attachments belong
to the Chinese QA group; English queries use translated text with unchanged
scene audio, so language-group differences do not isolate language alone.

Short audio chunks are padded with silence where required by the model's
processor, without reading future audio or advancing the observation clock.
Adapter-specific thresholds and causal-audio execution records are included
in the reproducibility materials.

\paragraph{AURA adaptation.}
Our AURA \citep{lu2026aura} adapter uses the official checkpoint with up to 30
recent one-second video segments (151,200 pixels per frame), reattaches the
request at each decision, and uses silent placeholders for earlier assistant
turns. It receives no raw scene audio and uses annotated query text for
assistance, cached ASR for historical audio-only queries, and original image
attachments.

\subsection{Model sizes and generation settings}
MiniCPM-o 4.5 and MOSS-VL have 9B and 11.3B full-model
parameters, respectively, despite both using a Qwen3-8B language backbone.
Proprietary parameter counts are undisclosed.

Local prompted models use greedy decoding; hosted calls use temperature zero.
Gemini-3.1-Pro and Gemini-3.5-Flash retain provider-default thinking, so their limits
are not answer-only token budgets. MiniCPM-o 4.5 uses greedy duplex decoding, 20 spoken tokens per chunk,
and a 30-unit context window. LiveCC uses a repetition penalty of 1.05 and resets
at 28,000 tokens; MMDuet2 resets at 100,000. MOSS-VL uses greedy session generation
with a 512-token setting and a pace of 12 tokens per second. AURA uses temperature
zero and a 512-token limit. Context retention and computation are not equalized
across interfaces. Recorded model revisions and configuration fingerprints are
retained in the reproducibility materials.
\subsection{Core prediction and scoring prompts}
\label{app:core-prompts}
The excerpts below document the instructions that determine when to respond and
what constitutes target coverage. Model-specific media serialization is described
above. Prompt wording is quoted from the recorded implementation; bracketed
field descriptions below denote instance-specific input. The metadata above each
divider describes usage and is not part of the quoted instruction.

\paragraph{Explicit-request decisions.}
Prompted T1--T7 adapters prepend the following instruction before the standing
request and the disclosed tool block:
\begin{promptbox}{T1--T7: response decision}{Prompt}
\promptmeta{Prompted decisions under an explicit request.}
{Causal scene prefix, standing request, and public tool definitions.}
{ABSTAIN, a response, or an inline tool call.}
You are a proactive, always-on streaming video assistant. You are shown video
frames sampled up to the CURRENT moment, plus the user's standing request.
Decide whether now is the right moment to respond.\par
- If the event/condition the user asked about has NOT happened yet, reply with
exactly: ABSTAIN\par
- If it HAS happened by now, give the appropriate response.
\end{promptbox}
The tool block supplies public names and definitions, the inline
\texttt{<call:NAME>} schema, and the instruction to use plain text when no tool
is warranted. It discloses neither target labels nor tool frequencies. Streaming
adapters retain their native standby mechanism while receiving the tool
disclosure. The request follows as \texttt{[Standing request]} and its text.

\paragraph{Assistance context and response policy.}
T8 supplies a user-profile sentence: either ``The user cannot see the scene.
Convey relevant visual information in words.'' or ``The user cannot hear the
scene. Convey relevant audible events and speech in text; use the original audio
and supporting visible clues.'' The requested language is then specified,
followed by:
\begin{promptbox}{T8: shared grounding instruction}{Prompt}
\promptmeta{Shared evidence constraint for T8-A/B/C.}
{Current causal scene evidence and the user profile/request.}
{Grounded assistance when a response is warranted.}
Use only the provided scene evidence up to now. Be accurate, concise and
actionable. Do not guess unseen objects, text, directions, distances or unheard
sounds.
\end{promptbox}
The subtask-specific sentence is:
\begin{promptbox}{T8-A/B/C: task instructions}{Prompt}
\promptmeta{Select the instruction for the evaluated assistance subtask.}
{User profile and scene evidence; T8-B/C also include the request.}
{A reminder, an answer, or the currently justified guidance step.}
\textbf{T8-A:} Monitor for useful assistance and timely safety reminders. State
what is happening and an appropriate action. Do not wait for a question. Avoid
redundant reminders.\par
\textbf{T8-B:} Answer the user's question once the evidence is sufficient.\par
\textbf{T8-C:} Guide the user's intention with short steps as the observed state
changes. Correct visible mistakes. Give only the step currently justified, not
future steps.
\end{promptbox}
Autonomous prompted calls end with ``If no useful new response is warranted now,
output exactly ABSTAIN.'' Native streaming uses ``Stay silent when there is no
useful new assistance to provide.'' T8-B/C append the user request; T8-A does not
append an event-specific query. In the paired event-guidance controls, a reviewed
event instruction is appended. Under either guidance condition, scheduled calls
request an answer at an annotated valid time with the sentence:
\begin{promptbox}{Scheduled response condition}{Prompt}
\promptmeta{Scheduled arms of the paired T8-A guidance control.}
{Current evidence at an annotated response opportunity.}
{One assistance response; the call supplies timing, not reference content.}
An external scheduler requests an answer now. Give the most useful assistance
supported by the current evidence. The scheduler does not reveal an event or answer.
\end{promptbox}
This wording describes the model-facing input; the experiment still supplies
temporal information through the scheduled call itself.

\paragraph{Assistance content judge.}
The recorded scoring prompt below evaluates reference-target coverage.
\begin{promptbox}{Assistance content judge}{Judge}
\promptmeta{Target-specific content scoring for T8.}
{User profile/request, reference target, and prediction; no media or time.}
{Exactly Correct or Incorrect; the scorer handles time and assignment.}
Judge the content of one accessibility assistance response.\par
The reference is an annotated target, not the only valid thing to say in the scene.\par
Correct means the prediction communicates this same target information and any essential action
without contradictory directions, unsafe advice, incorrect facts or invented critical details.\par
Equivalent wording is fine. A generic warning alone does not cover a specific target. A C step
must provide the reference's currently actionable guidance, not just repeat the user's intention.\par
Ignore timing (scored separately). Do not penalize a different language here: report language
compliance separately. If the prediction concerns another event, mark Incorrect FOR THIS TARGET;
that does not mean it is an invalid reminder. Do not infer validity of unannotated events.\par
User profile/request: \{question\}\par
Reference target: \{gt\}\par
Prediction: \{pred\}\par
Return exactly Correct or Incorrect.\par
\end{promptbox}
The prompt's instruction about language does not add a language-compliance
metric to the reported tables. Its verdict concerns semantic correspondence to
the supplied target; temporal eligibility and ordered assignment are handled
separately by the scorer. General, event, and translation criteria are summarized
in Table~\ref{tab:content-rubrics}.

\section{Additional Results and Diagnostics}
\label{app:full-diagnostics}
The analyses below expand the three main questions: timely action, query-free
initiative, and sustained coverage. Evaluation sensitivity and failure analyses
follow these results. Paired confidence intervals (CIs) follow
Appendix~\ref{app:uncertainty}.

\subsection{Timely action: tool choice and event timing}
\label{app:tool-decomposition}
Table~\ref{tab:t1-full} separates on-time first responses, joint tool-invocation
accuracy, and tool accuracy conditional on timing. If $T$ denotes an on-time
first response and $U$ the correct tool set, then
$P(T\cap U)=P(T)P(U\mid T)$. Joint and on-time rates use all 200 targets;
conditional accuracy uses each model's own timely subset. Per-tool scores retain
119 image-search, 69 web-search, and 12 audio-search targets.

\begin{table}[!htbp]\centering
\caption{T1 timing and tool accuracy (\%). Joint and on-time scores use 200 targets; tool$\mid$time uses each model's timely responses. Per-tool denominators are 119 image, 69 web, and 12 audio targets. Incomplete targets remain in the denominator.}
\label{tab:t1-full}
\begingroup\fontsize{9}{10.2}\selectfont\setlength{\tabcolsep}{3pt}
\begin{tabular*}{\linewidth}{@{\extracolsep{\fill}}lrrrrrr@{}}
\toprule
\textbf{Model} & \textbf{On-time} & \textbf{Joint} & \textbf{Tool$\mid$time} & \textbf{Image} & \textbf{Web} & \textbf{Audio} \\
\midrule
Gemini-3.1-Pro & 38.00 & 22.50 & 59.21 & 19.33 & 24.64 & 41.67 \\
Gemini-3.5-Flash & 37.50 & 29.50 & 78.67 & 21.85 & 42.03 & 33.33 \\
Qwen3.5-Omni-Plus & 39.00 & 32.00 & 82.05 & 26.89 & 39.13 & 41.67 \\
Qwen3-Omni & 37.00 & 26.50 & 71.62 & 18.49 & 37.68 & 41.67 \\
Qwen2.5-Omni & 28.00 & 11.00 & 39.29 & 7.56 & 15.94 & 16.67 \\
Qwen3-VL & 34.00 & 19.50 & 57.35 & 12.61 & 30.43 & 25.00 \\
InternVL3.5 & 38.00 & 29.00 & 76.32 & 23.53 & 39.13 & 25.00 \\
LiveCC & 40.00 & 0.00 & 0.00 & 0.00 & 0.00 & 0.00 \\
MiniCPM-o 4.5 & 6.00 & 1.50 & 25.00 & 0.00 & 2.90 & 8.33 \\
MMDuet2 & 14.00 & 0.00 & 0.00 & 0.00 & 0.00 & 0.00 \\
MOSS-VL & 17.50 & 1.50 & 8.57 & 0.84 & 2.90 & 0.00 \\
AURA & 37.50 & 0.00 & 0.00 & 0.00 & 0.00 & 0.00 \\
\bottomrule
\end{tabular*}
\endgroup
\end{table}

The full table also reveals failures in tool selection among timely streaming
responses: MOSS-VL selects the correct set in three of 35 timely responses,
whereas LiveCC has 80 timely responses and no joint successes. These are outcomes
of the evaluated model--adapter configurations.

\subsubsection{Request and timing priors}
\label{app:query-only}
\label{app:temporal-priors}
The request-only control supplies the same 200 request texts and public tool
definitions to Gemini-3.5-Flash, Qwen3-VL, and Qwen2.5-Omni, excluding scene media,
attachments, target times, answers, and interaction identifiers. It asks for image
search, web search, audio search, or abstention if the text is insufficient.
All models use a 2,048-token limit, with temperature zero for Gemini-3.5-Flash and greedy
local decoding. Exact-label scoring counts abstention or invalid labels as
incorrect. Gemini-3.5-Flash recovers 193/200 tool labels from the request alone
(Table~\ref{tab:query-only}a); this does not identify the search referent in the
stream.

We then emit either the majority tool (image search) or each model's frozen
request-only choice at four fixed delays after query arrival. Policies use only
query time and clip duration, withholding abstentions and calls beyond clip end.
Table~\ref{tab:query-fixed-time}b reports all four delays with the same 200-target
denominator and two-second scoring tolerance. Gemini-3.5-Flash reaches 68.00\% at three
seconds, the descriptive maximum among these delays. No independent development
set selected this delay, and the closed-label instruction and response
opportunities differ from the main evaluation. These controls diagnose request
and timing priors rather than provide a matched model ranking.

\begin{table}[!htbp]
\centering
\caption{Request and timing priors on 200 T1 targets, without scene input. (a) Request-only tool accuracy and class recalls (\%); class sizes are 119/69/12. Mean recall weights classes equally, and abstentions count as errors. (b) Joint hits (\%) when each tool policy is emitted at a fixed delay after query arrival. All tested delays are shown; their maximum was not independently selected.}
\label{tab:query-only}
\label{tab:query-fixed-time}
\begingroup\fontsize{9}{10.2}\selectfont
\setlength{\tabcolsep}{3pt}\renewcommand{\arraystretch}{1.10}
\par\noindent\textbf{(a) Tool choice from the request}\par\smallskip
\begin{tabular*}{\linewidth}{@{\extracolsep{\fill}}lrrrrrr@{}}
\toprule
\textbf{Predictor} & \textbf{Accuracy} & \textbf{Mean recall} & \textbf{Image} & \textbf{Web} & \textbf{Audio} & \textbf{Abstain} \\
\midrule
Always image search & 59.50 & 33.33 & 100.00 & 0.00 & 0.00 & 0 \\
Gemini-3.5-Flash & 96.50 & 97.43 & 96.64 & 95.65 & 100.00 & 4 \\
Qwen3-VL & 51.50 & 65.13 & 21.85 & 98.55 & 75.00 & 24 \\
Qwen2.5-Omni & 63.00 & 69.28 & 41.18 & 100.00 & 66.67 & 0 \\
\bottomrule
\end{tabular*}
\par\medskip
\par\noindent\textbf{(b) Tool choice plus a fixed response delay}\par\smallskip
\begin{tabular*}{\linewidth}{@{\extracolsep{\fill}}lrrrr@{}}
\toprule
\textbf{Tool-choice policy} & \textbf{$+0$ s} & \textbf{$+1$ s} & \textbf{$+3$ s} & \textbf{$+5$ s} \\
\midrule
Always image search & 40 (20.00) & 67 (33.50) & 83 (41.50) & 44 (22.00) \\
Gemini-3.5-Flash & 72 (36.00) & 107 (53.50) & 136 (68.00) & 72 (36.00) \\
Qwen3-VL & 43 (21.50) & 56 (28.00) & 72 (36.00) & 39 (19.50) \\
Qwen2.5-Omni & 49 (24.50) & 70 (35.00) & 90 (45.00) & 47 (23.50) \\
\bottomrule
\end{tabular*}
\endgroup
\end{table}

\subsubsection{Paired scene ablation}
\label{app:scene-control}
Gemini-3.5-Flash is rerun on all 200 T1 interactions with and without scene input. Scene
removal excludes video, audio, and transcripts while retaining the request,
its attachments, and tool definitions. The decision grid, prefix sampling,
decoding, and first-response scorer are identical. Execution stops at the first
response, which determines each single-target score; nonresponding runs cover the
full grid. These stopped traces do not measure full-clip output volume.

Scene-present attains 59 joint hits and 76 on-time responses, compared with 66
and 71 for scene-removed. Both have 123 premature responses, with one versus six
late or missing responses. Paired successes are
57 shared, two scene-present only, and nine scene-removed only. The joint-score
difference is $-3.50$ points (95\% CI: $[-7.00,-0.50]$, 200 clusters).
Both arms are separate runs. The main Gemini-3.5-Flash row has 59 hits and 75
on-time responses on the same 200 targets. The ablation measures input-removal effects,
without establishing counterfactual tracking of event occurrence or timing.

\subsection{Query-free initiative: reminder coverage and response policy}
\label{app:reminder-controls}
\subsubsection{Event guidance and scheduled responses}
\label{app:paired}
The paired experiment uses the same 122 T8-A episodes for Gemini-3.5-Flash, Qwen3-Omni,
and Qwen2.5-Omni. It crosses event-specific (E) or general (G) instructions with
scheduled (S) or autonomous (A) responding. Scheduled conditions request one
response at an annotated valid time. Autonomous conditions use the full causal
decision grid without previous responses in the prompt.

For episode $i$, let $Y_i(g,s)$ indicate a timely target match. Event-guidance
effects are $\Delta_s=122^{-1}\sum_i[Y_i(E,s)-Y_i(G,s)]$; the scheduling
interaction is $\Delta_S-\Delta_A$. Table~\ref{tab:paired-full} places the four
condition results alongside guidance effects and their interaction, using
114 video-record clusters.
Only Qwen2.5-Omni shows a statistically supported guidance-by-scheduling
interaction.
Three target windows contain no autonomous decision time. Restricting to the
119 reachable episodes (111 clusters) preserves every hit count except
Qwen2.5-Omni's event-scheduled count, which falls from eight to seven.
The reachable-subset interaction is $+6.72$ points for Qwen2.5-Omni
(95\% CI: $[+1.71,+12.00]$). Gemini-3.5-Flash's interval is
$[-2.52,+16.10]$ and Qwen3-Omni's is $[-4.17,+7.32]$, both including zero.
Scheduling changes temporal information and
response opportunities together, so it does not isolate a planning mechanism.

\begin{table}[!htbp]
\centering
\caption{Event guidance and response scheduling on the same 122 T8-A episodes. (a) Hits (recall, \%) in the four conditions. Scheduled (S) supplies an annotated response opportunity; Autonomous (A) uses the full causal grid. (b) Event-minus-general effects and their S-minus-A interaction in percentage points, with 95\% paired intervals from 114 video-record clusters.}
\label{tab:factorial}
\label{tab:paired-full}
\begingroup\fontsize{9}{10.2}\selectfont
\setlength{\tabcolsep}{3pt}\renewcommand{\arraystretch}{1.10}
\par\noindent\textbf{(a) Condition results}\par\smallskip
\begin{tabular*}{\linewidth}{@{\extracolsep{\fill}}lrrrr@{}}
\toprule
 & \multicolumn{2}{c}{\textbf{Event}} & \multicolumn{2}{c}{\textbf{General}} \\
\cmidrule(lr){2-3}\cmidrule(l){4-5}
\textbf{Model} & \textbf{Scheduled} & \textbf{Autonomous} & \textbf{Scheduled} & \textbf{Autonomous} \\
\midrule
Gemini-3.5-Flash & 30 (24.59) & 30 (24.59) & 19 (15.57) & 27 (22.13) \\
Qwen2.5-Omni & 8 (6.56) & 2 (1.64) & 0 (0.00) & 3 (2.46) \\
Qwen3-Omni & 8 (6.56) & 5 (4.10) & 4 (3.28) & 3 (2.46) \\
\bottomrule
\end{tabular*}
\par\medskip
\par\noindent\textbf{(b) Paired guidance effects [95\% CI]}\par\smallskip
\begin{tabular*}{\linewidth}{@{\extracolsep{\fill}}lrrr@{}}
\toprule
\textbf{Model} & \textbf{Event effect, S} & \textbf{Event effect, A} & \textbf{Interaction} \\
\midrule
Gemini-3.5-Flash & +9.02 [+0.82, +17.36] & +2.46 [-5.83, +11.21] & +6.56 [-2.52, +15.97] \\
Qwen2.5-Omni & +6.56 [+2.48, +11.29] & -0.82 [-4.27, +2.52] & +7.38 [+2.46, +12.80] \\
Qwen3-Omni & +3.28 [-1.65, +8.26] & +1.64 [-2.42, +5.83] & +1.64 [-4.07, +7.20] \\
\bottomrule
\end{tabular*}
\endgroup
\end{table}

\subsubsection{Response history and response cooldown}
\label{app:history-controls}
Three new Gemini-3.5-Flash runs use all 122 episodes under general assistance instructions:
no history, the five most recent timestamped responses, and the same history with
a 10-second cooldown between responses, measured in media time.
Each arm receives only its own actual response history,
without reference answers or match feedback. Other inputs, decision times, and
decoding settings are fixed. The model is informed of the cooldown rule.
After a response, an external controller skips generation calls for ten seconds.
Later decisions receive the current causal prefix. Abstention does not start a
cooldown. The intervention changes both output opportunities and
call count, rather than holding compute fixed.

\begin{table}[!htbp]
\centering
\caption{Gemini-3.5-Flash history and generation-time cooldown controls on 122 T8-A episodes. History contains the five most recent timestamped responses; cooldown enforces ten seconds between outputs. Paired intervals use 114 video-record clusters.}
\label{tab:history-cooldown}
\begingroup
\fontsize{9}{10.2}\selectfont
\setlength{\tabcolsep}{3pt}
\renewcommand{\arraystretch}{1.12}
\begin{tabular*}{\linewidth}{@{\extracolsep{\fill}}lrrr@{}}
\toprule
\textbf{Condition} & \textbf{Hits (recall, \%)} & \textbf{Outputs} & \textbf{Outputs/min} \\
\midrule
No history & 22 (18.03) & 7,833 & 58.39 \\
History & 22 (18.03) & 3,283 & 24.47 \\
History + cooldown & 10 (8.20) & 808 & 6.02 \\
\midrule
\multicolumn{4}{l}{\textbf{Paired recall changes (percentage points)}} \\
\multicolumn{2}{l}{\textbf{Contrast}} & \textbf{Difference} & \textbf{95\% CI} \\
\multicolumn{2}{l}{History $-$ no history} & $+0.00$ & $[-8.13,+8.00]$ \\
\multicolumn{2}{l}{History + cooldown $-$ history} & $-9.84$ & $[-17.24,-2.46]$ \\
\bottomrule
\end{tabular*}
\endgroup
\end{table}

History reduces outputs by 58.09\% with the same aggregate coverage of 22/122,
but only ten targets succeed in both arms, with 12 gained and 12 lost.
The difference interval does not establish equivalence. Adding cooldown reduces
coverage to 10/122 and outputs to 808.

These three conditions do not estimate a history-by-cooldown interaction because
a no-history-plus-cooldown arm is absent.
Reduced volume alone also does not establish reduced semantic redundancy.

These runs are distinct from Table~\ref{tab:assistance} and the guidance experiment:
the regenerated baseline has 22 hits and 7,833 outputs, compared with the earlier
Gemini-3.5-Flash general condition's 27 hits and 7,813 outputs. All contrasts in
Table~\ref{tab:history-cooldown} use the regenerated arms and 114 paired
video-record clusters. These findings concern Gemini-3.5-Flash under the specified policies.

\subsubsection{Filtering recorded outputs}
\label{app:budgets}
We also apply first-$K$ and minimum-spacing filters to the earlier autonomous
trajectories. First-$K$ retains the first 1, 3, 5, or 10
responses per episode; spacing filters retain the first response and subsequent
responses at least 5 or 10 seconds after the last retained one. Filtering uses
no reference times, and matching is recomputed on retained outputs.

Under general instructions, five-second spacing leaves Gemini-3.5-Flash with
11 of 27 target matches and 1,642 of 7,813 outputs.
For Qwen3-Omni, the same spacing leaves two of three matches and 1,666 of 8,213
outputs. Unlike the response cooldown, these filters cannot change subsequent
model behavior. Both analyses report coverage together with output volume;
unmatched outputs are not automatically false positives
(Appendix~\ref{app:scoring}).

\subsection{Sustained coverage: steps and complete guidance}
\label{app:sequences}
The 26 T8-C interactions contain 20 three-step sequences, four four-step
sequences, and one each with five and six steps. Table~\ref{tab:assistance}
reports overall step and complete coverage. We examine the largest length group
below. 
The four-, five-, and six-step groups are too small to establish a length effect.

To interpret complete coverage, we independently permute the binary step-success
outcomes across episodes of the same length within each step position. For $n$ episodes of length
$K$ and marginal success rates $p_k$, the expected complete count is
\begin{equation}
\mathbb{E}_{\mathrm{perm}}[N_{\mathrm{complete}}]=n\prod_{k=1}^{K}p_k.
\end{equation}
Figure~\ref{fig:assistance} shows all models on the 20 three-step episodes, with
95\% reference intervals from 10,000 permutations. These intervals
describe the permutation distribution, not uncertainty in the observed rate.
The reference assumes exchangeability across episodes within length and does not control
shared episode difficulty.

\begin{appendixfigure}
\centering
\includegraphics[width=\linewidth]{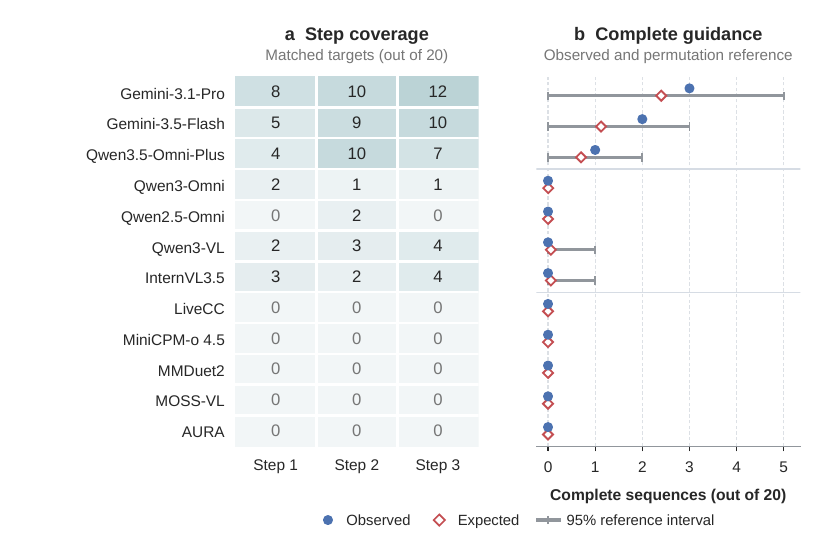}
\caption{\textbf{Step coverage and complete guidance on 20 three-step episodes.}
(a) Matched targets at each step position, with a denominator of 20 per cell.
(b) Observed complete counts, permutation expectations, and central 95\%
reference intervals. Each permutation preserves the marginal counts in (a).
The reference does not model episode difficulty or identify a memory mechanism.}
\label{fig:assistance}
\end{appendixfigure}

For Gemini-3.1-Pro, the step counts are 8/20, 10/20, and 12/20, totaling 30/60 covered steps. Its expected complete count
is $20\times0.4\times0.5\times0.6=2.40$, with a reference interval of $[0,5]$
containing the observed three completions. Gemini-3.5-Flash's observed two completions
likewise lie within $[0,3]$ around an expectation of 1.125. These results motivate
reporting complete guidance alongside step coverage, without establishing an
additional continuity deficit beyond the step-level outcomes. Both measures
concern the recorded reference trajectory, not closed-loop task success.

\subsection{Evaluation sensitivity and failure analysis}
\subsubsection{Decision schedules}
\label{app:protocol-control}
After release filtering, the cohort selected before inference retains 19 T1
interactions and 20 each from T2--T4, T6, and T7: 119 interactions, 142 targets,
and 117 video-record clusters.
Gemini-3.5-Flash and Qwen3-VL each run two schedules with the same query preparation,
historical context, frame sampling, modality, decoding, and judge. Target-local
probing uses one-second decisions within four seconds of each reference time,
clipped to query and media boundaries. Continuous probing uses the main
query-plus-integer grid. Thus both the extent and phase of the schedule change.

\begin{table}[!htbp]\centering
\caption{Decision-schedule effects on 119 paired interactions (142 targets). Accuracy equally weights T1--T4, T6, and T7. Differences are continuous minus target-local, in percentage points, with 95\% paired intervals.}
\label{tab:protocol-effects}
\begingroup\fontsize{9}{10.2}\selectfont\setlength{\tabcolsep}{3pt}
\begin{tabular*}{\linewidth}{@{\extracolsep{\fill}}lrrr@{}}
\toprule
\textbf{Model} & \textbf{Target-local} & \textbf{Continuous} & \textbf{Difference [95\% CI]} \\
\midrule
Gemini-3.5-Flash & 26.80 & 19.17 & $-7.63\;[-13.96,-1.77]$ \\
Qwen3-VL & 4.28 & 3.36 & $-0.92\;[-4.17,+2.10]$ \\
\bottomrule
\end{tabular*}
\endgroup
\end{table}

Table~\ref{tab:protocol-effects} reports equally weighted six-task means. Gemini-3.5-Flash's
accuracy falls by 7.63 points as premature responses rise from 58.45\% to 66.90\%
of targets. First-response scoring exposes early actions that target-local
probing cannot observe. Qwen3-VL's change has an interval spanning zero.
Premature fractions weight targets equally; these within-model contrasts do not
isolate architecture or modality effects.

\subsubsection{Temporal-tolerance sensitivity}
\label{app:t1-tolerance}
We rescore the same 400 trajectories at one-, two-, and four-second tolerances,
keeping predictions, tool extraction, first-response assignment, and the
separate two-second output-grouping rule fixed. Table~\ref{tab:t1-tolerance}
shows that absolute scores change substantially. All scene-present minus
scene-removed estimates are negative, but the one-second interval includes zero.
The absence of a positive point estimate persists across these windows; neither
score magnitude nor statistical support is invariant. The main tolerance remains
two seconds, and this sensitivity result is specific to the Gemini-3.5-Flash control.

\begin{table}[!htbp]
\centering
\caption{Temporal-tolerance sensitivity of the Gemini-3.5-Flash scene control. Cells give joint accuracy (\%) and hits out of 200. Differences are scene present minus removed, in percentage points, with 95\% paired intervals. The output-grouping rule remains fixed at two seconds.}
\label{tab:t1-tolerance}
\begingroup
\fontsize{9}{10.2}\selectfont
\setlength{\tabcolsep}{3pt}
\renewcommand{\arraystretch}{1.12}
\begin{tabular*}{\linewidth}{@{\extracolsep{\fill}}lrrr@{}}
\toprule
\textbf{Tolerance (s)} & \textbf{Scene present} & \textbf{Scene removed} & \textbf{Difference [95\% CI]} \\
\midrule
1 & 11.00 (22/200) & 12.50 (25/200) & $-1.50\;[-4.00,+1.00]$ \\
2 (primary) & 29.50 (59/200) & 33.00 (66/200) & $-3.50\;[-7.00,-0.50]$ \\
4 & 53.00 (106/200) & 66.00 (132/200) & $-13.00\;[-18.50,-7.50]$ \\
\bottomrule
\end{tabular*}
\endgroup
\end{table}

\subsubsection{Timing and content failures}
\label{app:failures}
Table~\ref{tab:failure-breakdown} decomposes the main results for state tracking
(T4) and continuous responses (T7). Gemini-3.5-Flash's T4 failures are predominantly
premature (172/206 targets), while T7 has both premature responses (233/457) and
timely content failures (152/457). A premature response fails regardless of its
content, whose correctness cannot be inferred from this breakdown.

\begin{table}[!htbp]
\centering
\caption{First-response outcomes on state tracking (T4) and continuous responses (T7), using the main-table trajectories. Counts sum to the target denominator in each row. Content failure denotes an on-time response with incorrect content.}
\label{tab:failure-breakdown}
\begingroup
\fontsize{9}{10.2}\selectfont
\setlength{\tabcolsep}{3pt}
\renewcommand{\arraystretch}{1.12}
\begin{tabular*}{\linewidth}{@{\extracolsep{\fill}}llrrrrr@{}}
\toprule
\textbf{Model} & \textbf{Task} & \textbf{Targets} & \textbf{Success} & \textbf{Premature} & \textbf{Content fail} & \textbf{Late/missing} \\
\midrule
Gemini-3.1-Pro & T4 & 206 & 23 & 178 & 5 & 0 \\
Gemini-3.1-Pro & T7 & 457 & 51 & 229 & 146 & 31 \\
Gemini-3.5-Flash & T4 & 206 & 21 & 172 & 10 & 3 \\
Gemini-3.5-Flash & T7 & 457 & 52 & 233 & 152 & 20 \\
\bottomrule
\end{tabular*}
\endgroup
\end{table}

\subsubsection{Scene audio with the request held fixed}
\label{app:modality}
Excluding query-audio attachments and audio-type queries leaves 1,133 interactions
and 1,383 targets. Conditions share query text, target-local decisions, and context
policy within each model. V provides video, V+ASR adds transcripts, and V,A adds
native scene audio. These runs use a separate schedule and earlier verdict handling (Appendix~\ref{app:scoring}) from the main comparison.
\begin{appendixtable}
\centering
\caption{Scene-audio effects with the query held fixed under target-local probing. (a) Joint hit counts; each model uses the same target denominator. V: video, V+ASR: video and transcripts, V,A: video and native audio. (b) Native-audio contrasts in percentage points with 95\% paired intervals (47 auditory and 27 audiovisual video-record clusters). These runs use a separate schedule and earlier verdict handling from the main comparison.}
\label{tab:modality-subsets}
\label{tab:modality-contrasts}
\begingroup\fontsize{9}{10.2}\selectfont
\setlength{\tabcolsep}{3pt}\renewcommand{\arraystretch}{1.10}
\par\noindent\textbf{(a) Overall and audio-related target subsets}\par\smallskip
\begin{tabular*}{\linewidth}{@{\extracolsep{\fill}}llrrrrrrrrr@{}}
\toprule
 & & \multicolumn{3}{c}{\textbf{Gemini-3.5-Flash}} & \multicolumn{3}{c}{\textbf{Qwen3-Omni}} & \multicolumn{3}{c}{\textbf{Qwen2.5-Omni}} \\
\cmidrule(lr){3-5}\cmidrule(lr){6-8}\cmidrule(l){9-11}
\textbf{Subset} & \textbf{Targets} & \textbf{V} & \textbf{V+ASR} & \textbf{V,A} & \textbf{V} & \textbf{V+ASR} & \textbf{V,A} & \textbf{V} & \textbf{V+ASR} & \textbf{V,A} \\
\midrule
All with T5 & 1383 & 313 & 313 & 317 & 110 & 116 & 123 & 39 & 42 & 50 \\
\midrule
Auditory & 78 & 6 & 6 & 13 & 4 & 8 & 11 & 3 & 5 & 5 \\
Audiovisual & 35 & 4 & 4 & 5 & 0 & 2 & 1 & 0 & 1 & 0 \\
\bottomrule
\end{tabular*}
\par\medskip
\begin{tabular*}{\linewidth}{@{\extracolsep{\fill}}llrrr@{}}
\toprule
\multicolumn{5}{l}{\textbf{Paired scene-audio effects (percentage points)}} \\
\textbf{Trigger} & \textbf{Comparison} & \textbf{Gemini-3.5-Flash} & \textbf{Qwen3-Omni} & \textbf{Qwen2.5-Omni} \\
\midrule
Auditory & V, A $-$ V & $+8.97\,[+3.33,\,+15.71]$ & $+8.97\,[+3.41,\,+16.39]$ & $+2.56\,[+0.00,\,+7.04]$ \\
 & V, A $-$ V+ASR & $+8.97\,[+3.33,\,+15.79]$ & $+3.85\,[-2.60,\,+10.98]$ & $+0.00\,[-5.19,\,+5.26]$ \\
Audiovisual & V, A $-$ V & $+2.86\,[-9.76,\,+16.13]$ & $+2.86\,[+0.00,\,+9.68]$ & $+0.00\,[+0.00,\,+0.00]$ \\
 & V, A $-$ V+ASR & $+2.86\,[-9.68,\,+16.13]$ & $-2.86\,[-13.51,\,+6.45]$ & $-2.86\,[-9.09,\,+0.00]$ \\
\bottomrule
\end{tabular*}
\endgroup
\end{appendixtable}

Table~\ref{tab:modality-subsets} retains the overall result and the auditory and
audiovisual subsets. The complete five-way trigger partition and the aggregate
excluding T5 (1,268 targets) accompany the supplementary data.
These aggregates weight targets equally, unlike the main six-task Avg.
Auditory and audiovisual contrasts use 47 and 27 video-record clusters,
respectively.

Native audio adds seven hits on 78 auditory targets for both Gemini-3.5-Flash
(6 to 13) and Qwen3-Omni (4 to 11), gains of 8.97 points with intervals above zero.
Gemini-3.5-Flash gains only 0.29 points overall: aggregation obscures the
audio-specific benefit. These effects do not prove audio necessity for every target.

\begin{caseblock}
\section{Qualitative Cases}
\label{app:qualitative}
\subsection{Timely action and query-free reminders}
\label{app:case-tool}
\label{app:case-reminder}
Figure~\ref{fig:cases-action-initiative} pairs real source frames with reference
requirements and recorded model outputs. These selected cases illustrate failure
modes, not their prevalence. Times are rounded for display; verdicts are taken
from the stored scorer outputs.

\begin{appendixfigure}
\includegraphics[width=\linewidth]{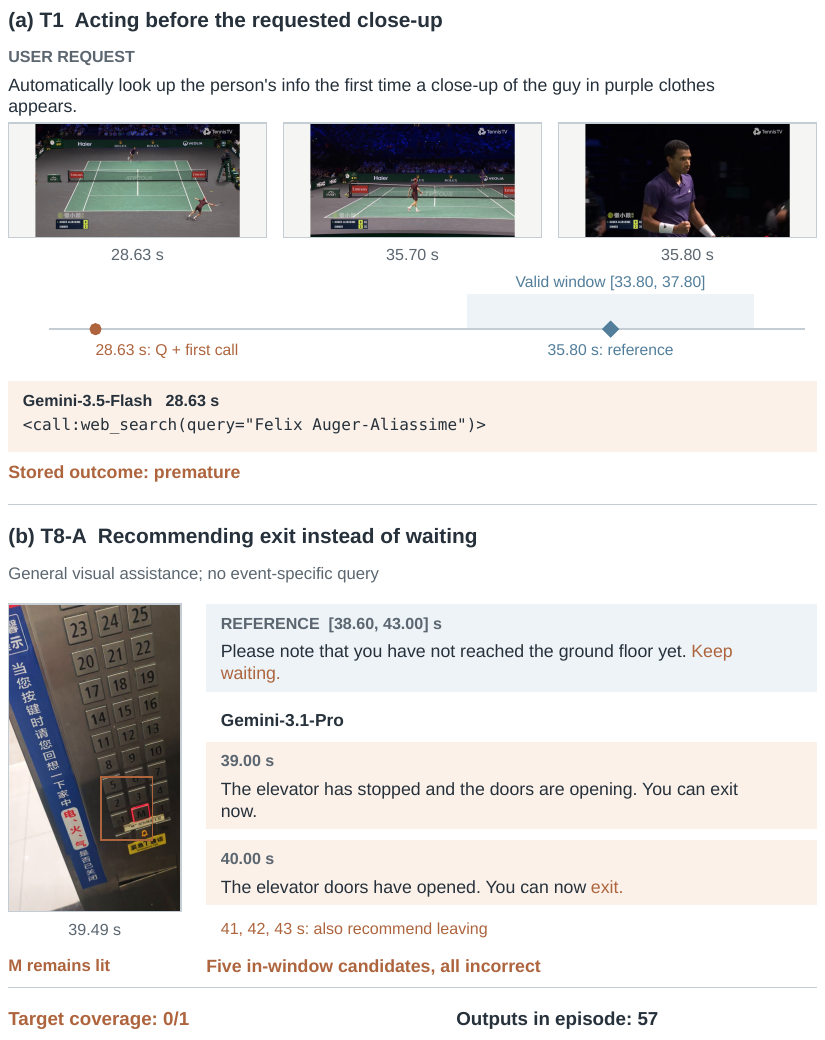}
\caption{\textbf{Acting early and giving the wrong reminder.}
(a) Gemini-3.5-Flash calls a tool at query arrival, before the scored window,
and repeats it at 29.00 s. The reference tool is \texttt{image\_search}; timing
already determines failure. (b) The still-lit M button is outlined in the source
frame. All five in-window Gemini-3.1-Pro responses recommend exit instead of
waiting. Two are quoted in full; other unmatched outputs are not thereby
classified as invalid assistance. Neither case establishes an internal cause.}
\label{fig:cases-action-initiative}
\end{appendixfigure}
\end{caseblock}

\begin{caseblock}
\subsection{Code diagnosis under changing screen states}
\label{app:case-code}
Figure~\ref{fig:case-code} contrasts locally grounded advice with a later
screen-inconsistent diagnosis. Gemini-3.1-Pro identifies a duplicate variable and
an extra semicolon, but then claims that the initializer braces have been deleted
when both are visible. Useful earlier advice therefore does not establish that
subsequent recommendations remain consistent with the current code.

\begin{appendixfigure}
\includegraphics[width=\linewidth]{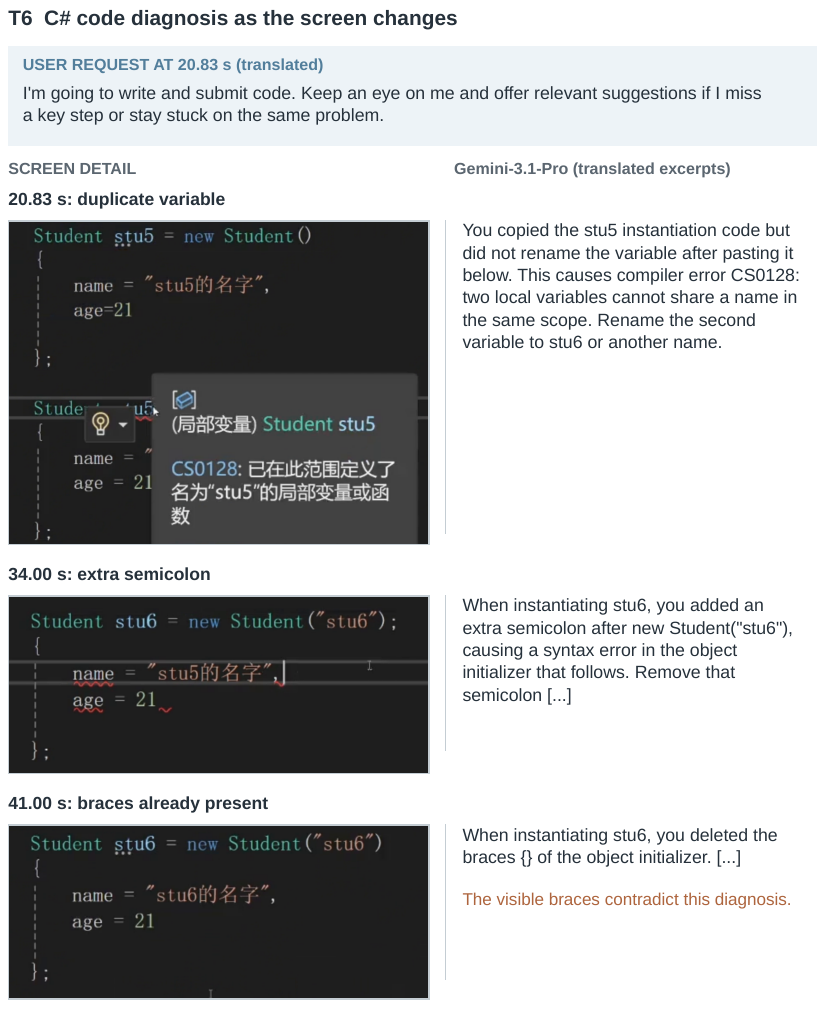}
\caption{\textbf{C\# diagnosis across three editing states.}
Source-frame crops accompany recorded replies at the indicated times. The request
and replies are translated from Chinese; \texttt{[...]} marks omissions.
The original target is a reminder to run tests at 60.50 s, against which the stored
outcome is premature. The displayed diagnoses illustrate agreement with the
screen, not additional scored targets.}
\label{fig:case-code}
\end{appendixfigure}
\end{caseblock}

\stepcounter{footnote}
\begin{caseblock}
\subsection{A tutorial reminder at the requested stage}
\label{app:case-import}
In Figure~\ref{fig:case-import}, the user requests a reminder when a tutorial
introduces its second media-import method. Qwen3-Omni responds at 162.00 s, before
the reference window. Gemini-3.1-Pro first responds at 191.00 s and receives a
stored match. The contrast illustrates why a plausible description of an
operation must also be tied to the requested stage of the demonstration.

\begin{appendixfigure}
\includegraphics[width=\linewidth]{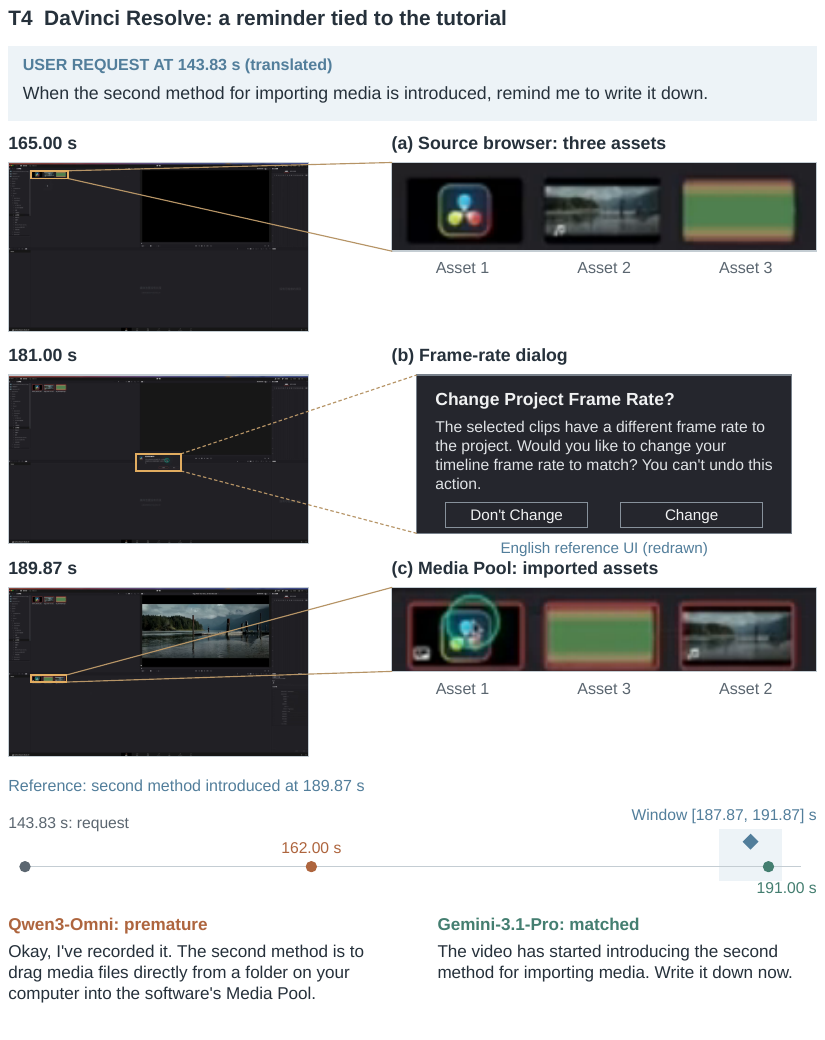}
\vspace{-14pt}
\caption{\textbf{Following a DaVinci Resolve import workflow.}
Original frames appear on the left. Panels (a,c) show source crops with matching
asset labels; (b) is a redrawn English reference dialog.\protect\footnotemark[\value{footnote}]
The diamond marks the annotated second-method point (189.87 s), with its
$\pm2$-second window shaded. Circles mark the request and first replies.}
\label{fig:case-import}
\end{appendixfigure}
\end{caseblock}
\unskip\nopagebreak
\footnotetext[\value{footnote}]{Chris Roberts and Simon Hall, \emph{The Beginner's Guide to DaVinci
Resolve 20}, Blackmagic Design, 2025, p.~351.
\href{https://documents.blackmagicdesign.com/UserManuals/DaVinci-Resolve-20-Beginners-Guide.pdf}{Official PDF}.}
\begin{caseblock}
\subsection{Partial coverage within a guidance sequence}
\label{app:case-guidance}
Figure~\ref{fig:case-guidance} shows two matched steps and a missed final target.
All four final-target candidates at 32--35 seconds are incorrect, giving $2/3$
target coverage and zero complete sequences despite 67 outputs. This illustrates
partial coverage without identifying a memory or planning mechanism.

\begin{appendixfigure}
\includegraphics[width=\linewidth]{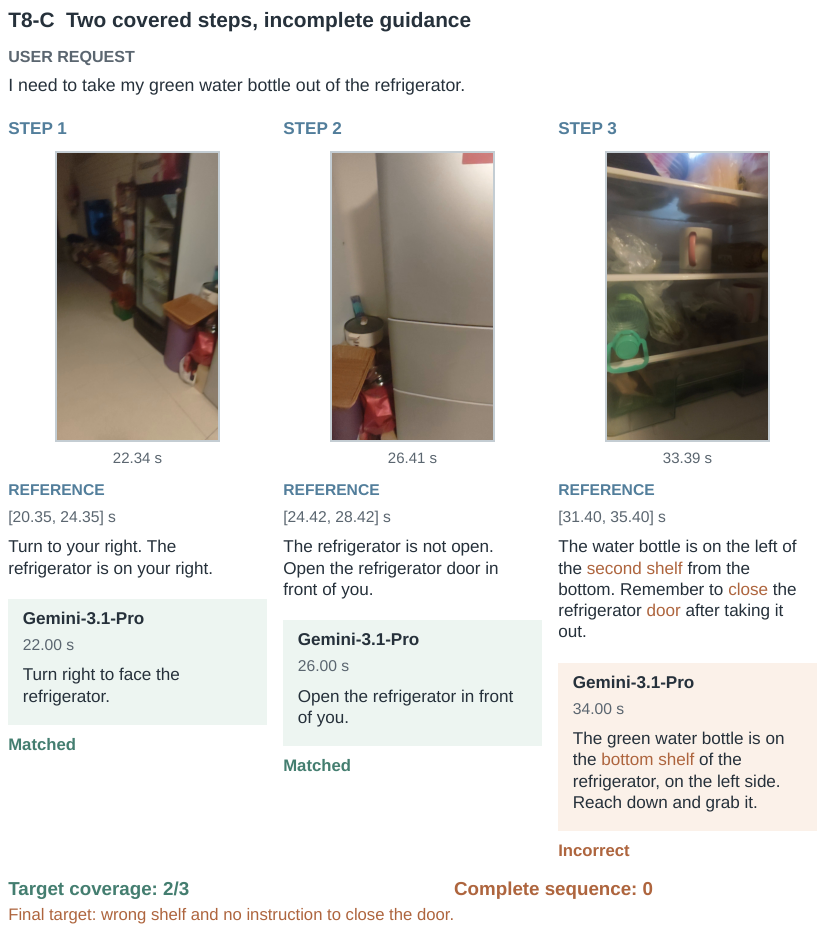}
\caption{\textbf{Two covered steps with an incomplete final instruction.}
Each column pairs a source state with the full reference and a recorded
Gemini-3.1-Pro response. The first two columns show stored matches; the third
shows one of four incorrect final-target candidates. The shelf location and
closing action distinguish the reference from the prediction. Frame timestamps
identify sampled evidence, while output timestamps identify model emissions.
An incorrect 21-second attempt does not prevent the later 22-second match under
ordered assignment.}
\label{fig:case-guidance}
\end{appendixfigure}
\end{caseblock}

\end{document}